\documentclass[letterpaper]{article} % DO NOT CHANGE THIS
\usepackage{aaai24}  % DO NOT CHANGE THIS
\usepackage{times}  % DO NOT CHANGE THIS
\usepackage{helvet}  % DO NOT CHANGE THIS
\usepackage{courier}  % DO NOT CHANGE THIS
\usepackage[hyphens]{url}  % DO NOT CHANGE THIS
\usepackage{graphicx} % DO NOT CHANGE THIS
\usepackage{natbib}  % DO NOT CHANGE THIS AND DO NOT ADD ANY OPTIONS TO IT
\usepackage{caption} % DO NOT CHANGE THIS AND DO NOT ADD ANY OPTIONS TO IT
\usepackage{algorithm}
\usepackage{algorithmic}

\usepackage{newfloat}
\usepackage{listings}
\DeclareCaptionStyle{ruled}{labelfont=normalfont,labelsep=colon,strut=off} % DO NOT CHANGE THIS
\floatstyle{ruled}
\newfloat{listing}{tb}{lst}{}
\floatname{listing}{Listing}
\def\ourmethod{Auto-Prox}
\usepackage{multirow}
\usepackage{booktabs}
\usepackage{amsmath}
\usepackage{amssymb}
\usepackage{dsfont}
\usepackage{color}
\usepackage{listings}
\definecolor{deepblue}{rgb}{0,0,0.5}
\definecolor{deepred}{rgb}{0.6,0,0}
\definecolor{deepgreen}{rgb}{0,0.5,0}
\definecolor{codebrown}{rgb}{0.8,0.44,0.2}
\definecolor{codegray}{rgb}{0.5,0.5,0.5}
\definecolor{codepurple}{rgb}{0.58,0,0.82}
\definecolor{backcolour}{rgb}{0.95,0.95,0.92}
\definecolor{codegreen}{rgb}{0,0.6,0}

\newcommand{\textcode}[1]{{\normalfont\codefont #1}}

\def\codefont{\fontfamily{lmtt}\selectfont}

\title{Auto-Prox: Training-Free Vision Transformer Architecture Search via Automatic Proxy Discovery}
\author{
    Zimian Wei\textsuperscript{\rm 1}\thanks{These authors contributed equally.},
    Lujun Li\textsuperscript{\rm 2}\footnotemark[1],
    Peijie Dong\textsuperscript{\rm 3},
    Zheng Hui\textsuperscript{\rm 4},
    Anggeng Li\textsuperscript{\rm 5},
    Menglong Lu\textsuperscript{\rm 1},
    Hengyue Pan\textsuperscript{\rm 1}\thanks{Corresponding author},
    Zhiliang Tian\textsuperscript{\rm 1},
    Dongsheng Li\textsuperscript{\rm 1}\footnotemark[2]
}
\affiliations{
    \textsuperscript{\rm 1}National University of Defense Technology, \textsuperscript{\rm 2}HKUST, \textsuperscript{\rm 3}HKUST(GZ),    \textsuperscript{\rm 4}Columbia University, \textsuperscript{\rm 5}Huawei\\

   \{weizimian16,lumenglong,hengyuepan,tianzhiliang, dsli\}@nudt.edu.cn,\{lilujunai,dongpeijie98\}@gmail.com\\
   
   Zh2483@columbia.edu,anggeng.li@outlook.com
}

\lstdefinestyle{mystyle}{
    backgroundcolor=\color{backcolour},   
    commentstyle=\color{codegreen},
    keywordstyle=\color{magenta},
    numberstyle=\tiny\color{codegray},
    stringstyle=\color{codepurple},
    basicstyle=\ttfamily\footnotesize,
    breakatwhitespace=false,         
    breaklines=true,                 
    captionpos=b,                    
    keepspaces=true,                 
    numbers=left,                    
    numbersep=5pt,                  
    showspaces=false,                
    showstringspaces=false,
    showtabs=false,                  
    tabsize=2
}

\newcommand{\tabincell}[2]{\begin{tabular}{@{}#1@{}}#2\end{tabular}}

\begin{document}

\maketitle

\begin{abstract}

% Vision Transformer (ViT) has achieved substantial success in various computer vision tasks. 
% Efficient neural architecture search for ViT is critical for its practical application. 
The substantial success of Vision Transformer (ViT) in computer vision tasks is largely attributed to the architecture design.
This underscores the necessity of efficient architecture search for designing better ViTs automatically.
As training-based architecture search methods are computationally intensive, there’s a growing interest in training-free methods that use zero-cost proxies to score ViTs. However, existing training-free approaches require expert knowledge to manually design specific zero-cost proxies. Moreover, these zero-cost proxies exhibit limitations to generalize across diverse domains. In this paper, we introduce Auto-Prox, an automatic proxy discovery framework, to address the problem. First, we build the ViT-Bench-101, which involves different ViT candidates and their actual performance on multiple datasets. Utilizing ViT-Bench-101, we can evaluate zero-cost proxies based on their score-accuracy correlation. Then, we represent zero-cost proxies with computation graphs and organize the zero-cost proxy search space with ViT statistics and primitive operations. To discover generic zero-cost proxies, we propose a joint correlation metric to evolve and mutate different zero-cost proxy candidates. We introduce an elitism-preserve strategy for search efficiency to achieve a better trade-off between exploitation and exploration. Based on the discovered zero-cost proxy, we conduct a ViT architecture search in a training-free manner. Extensive experiments demonstrate that our method generalizes well to different datasets and achieves state-of-the-art results both in ranking correlation and final accuracy. Codes can be found at https://github.com/lilujunai/Auto-Prox-AAAI24.
\end{abstract}

\section{Introduction}

Recently, Vision Transformer (ViT)~\cite{vit} has achieved remarkable performance in image classification~\cite{evit,lvvit,crossvit}, object detection~\cite{Wu2022TinyViTFP}, semantic segmentation~\cite{cswin}, and other computer vision tasks~\cite{localvit,swin}. 
Despite these advancements, the manual trial-and-error method of designing ViT architectures becomes impractical given the expanding neural architecture design spaces and intricate application scenarios~\cite{liu2023norm,li2023auto,li2022shadow,li2023kd,li2022tf,li2022SFF,li2022self,shao2023catch}.
Neural Architecture Search (NAS) aims to address this issue by automating the design of neural network architectures.
Traditional training-based architecture search methods~\cite{snas,wei2024TVT,li2021nas,linas2,lichengp,dong2023diswot,dong2023emq, lu2024uniads,wei2024auto}
involve training and evaluating numerous candidate ViTs, which can be
computationally expensive and time-consuming. 
Therefore, there is a need for a more efficient architecture search of ViT.

\begin{figure}[t]
    \centering
    \includegraphics[width=1.0\linewidth]{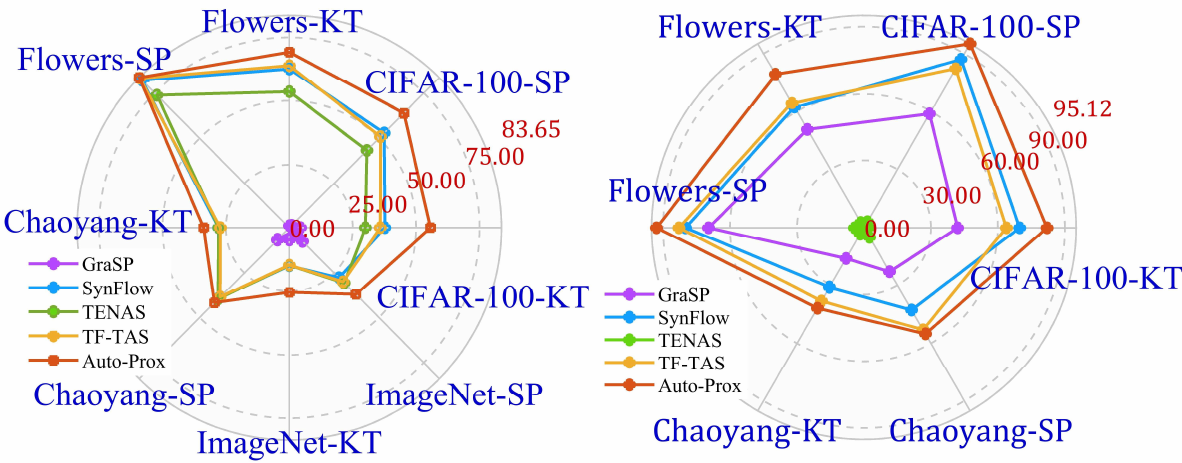}
    \caption{Kendall (KT) \& Spearman (SP) ranking correlations of zero-cost proxies on AutoFormer (Left)  and PiT (Right) search space for four datasets including CIFAR-100, Flowers, Chaoyang, and ImageNet. Results demonstrate that our proposed \ourmethod{} significantly outperforms Synflow~\protect\cite{syflow} and TF-TAS \protect \cite{DSS}.}
    \label{fig:accuracy}
       % \vspace{-1em}
\end{figure}

Recent training-free NAS methods, such as NWOT~\cite{NWOT} and TF-TAS \cite{DSS}, have received great research interest due to their meager costs. 
These methods utilize hand-crafted zero-cost proxies~\cite{syflow,TENAS}, which are conditional on the model's parameters or gradients, to predict the actual accuracy ranking without the expensive training process. 
However, there are still some drawbacks limiting their broader application: \textbf{(1) Dependency on expert knowledge and extensive tuning.} 
Lots of traditional zero-cost proxies are transferred from different areas with extensive expert intuition and time-consuming tuning processes. 
In addition, these hand-crafted zero-cost proxies can be influenced by human biases and limited by the designer's experience. 
\textbf{(2) Generality and flexibility.} Hand-crafted zero-cost proxies may perform well on the specific problem but can not generalize well to new or unseen datasets or tasks (see Figures~\ref{fig:accuracy}). These zero-cost proxies usually have fixed formulas, and some do not associate with input or labels of the target dataset. i.e., scores of the same architecture on different datasets are the same, which does not match the facts. Thus, two problems are naturally raised: \textbf{\textit{(1) How to efficiently discover the proxies without expert knowledge? (2) How to reduce the gap between fixed zero-cost proxy and variable tasks?}}

For the first problem, we propose Auto-Prox, a from-scratch automatic proxy search framework, as an alternative to traditional manual designs. 
Unlike hand-crafted zero-cost proxies, Auto-Prox mitigates human bias and automates the exploration of more expressive and efficient zero-cost proxies.
First, we establish the ViT-Bench-101 dataset, which comprises diverse ViT architectures and their corresponding performance on multiple datasets. ViT-Bench-101 provides a benchmark for evaluating the score-accuracy correlations of different zero-cost proxies. 
We then define the zero-cost proxy search space, which includes ViT's weights and gradient statistics as candidate inputs, potential unary and binary mathematical operations, and computation graphs as representations of zero-cost proxies. 
In our exploration of the proxy search space, we introduce an elitism preservation strategy to enhance search efficiency. 
This strategy involves judgment in the evolutionary search to preserve high-performing zero-cost proxies.

For the second problem, we propose a joint correlation metric, designed as the objective for the evolutionary proxy search.   
Instead of solely optimizing the automatic proxy based on high score-accuracy correlation within a single dataset, this metric captures the weighted average of correlations spanning multiple datasets. 
Based on the joint correlation metric, \ourmethod{} improves generalization, allowing for discovering zero-cost proxies that perform well on new or unseen datasets or tasks.

\begin{figure*}[hbtp]
    \centering
    \includegraphics[width=1.0\linewidth]{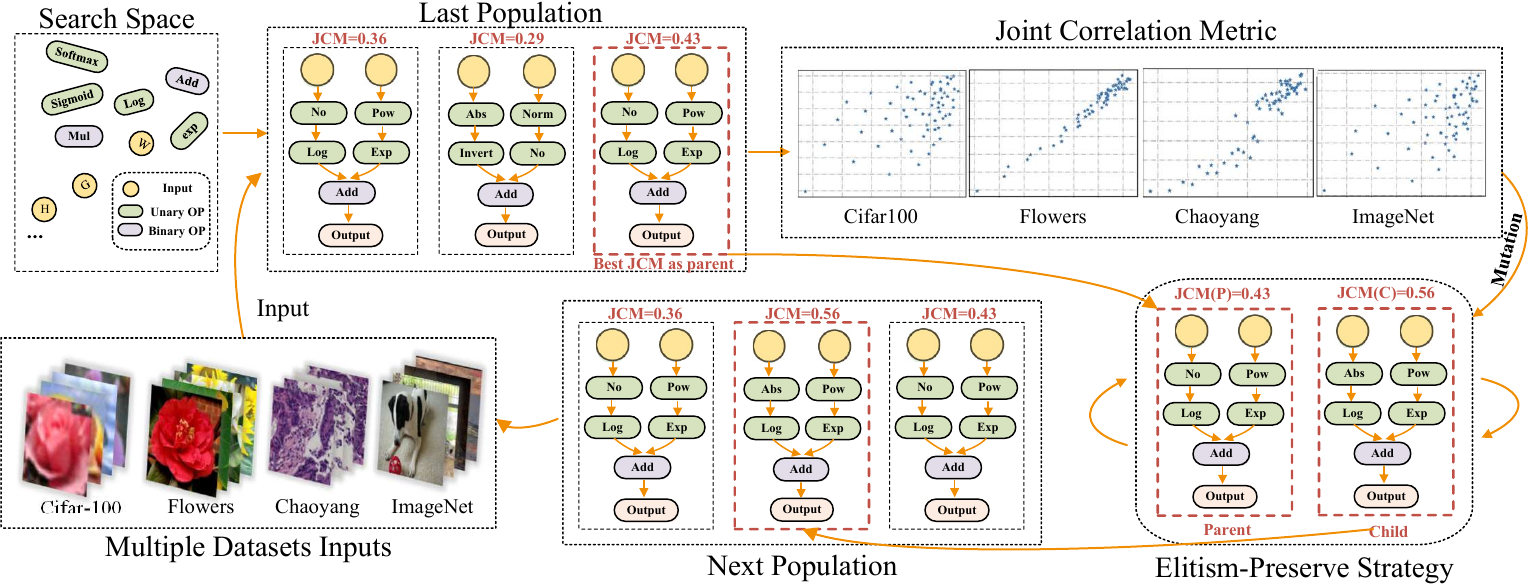}
    \caption{Illustration of the Auto-Prox search process.  First, we devise a comprehensive search space, incorporating primitive operations, ViT statistics, and computation graphs to represent zero-cost proxies. 
    We then randomly sample candidate zero-cost proxies to initialize the population and evaluate their ranking consistency using the Joint Correlation Metric across four datasets. 
    Based on the JCM score, we pick up promising ones as parents and perform mutation to generate a new population. 
    Subsequently, we perform the elitism-preserve strategy to prevent the deterioration of the population.  
    % Finally, we remove the lower-performing individual to keep the superiority of the population.
    }
    \label{fig:main}
\end{figure*}

We conduct extensive experiments on CIFAR-100, Flowers, Chaoyang~\cite{zhu2021hard}, and ImageNet-1K to validate the superiority of our proposed method. 
For small datasets, except for ImageNet-1K, we focus on distillation accuracy instead of vanilla accuracy for ViTs, in contrast to traditional NAS experiments.
The experiments demonstrate that our Auto-Prox can achieve better distillation accuracy than other zero-cost proxies when searched in the same search spaces.
Moreover, Auto-Prox obtains state-of-the-art ranking correlation across multiple datasets, significantly surpassing existing training-free NAS approaches (see Figure~\ref{fig:accuracy}) without prior knowledge.

% For efficiency, \ourmethod{} is free from back-propagation and can finish the entire search process within one GPU hour, significantly reducing the architecture search cost. 
% Our framework allows for the automatic discovery of more expressive, efficient, and effective proxies in optimizing and customizing ViT architectures, opening up new avenues for research and applications in ViT architecture search. 

% For example, \ourmethod{} can largely boost the ViT performance on tiny datasets (9.98\% for AutoFormer and 3.68\% for PiT on the CIFAR-100 dataset). 
% We also conduct comprehensive ablation studies to investigate how our method can leverage the predictability of zero-cost proxy to boost performance.
 
% For example, \ourmethod{} can largely boost the ViT performance on tiny datasets (e.g., 9.98\% for AutoFormer and 3.68\% for PiT on the CIFAR-100 dataset). 
% Additionally, our Auto-Prox obtains state-of-the-art results on rank consistency, such as 40\% Kendall correlation over TF-TAS~\cite{DSS}. For efficiency, \ourmethod{} is free from back-propagation and can finish the entire search process within one GPU-hour, significantly reducing the architecture search cost. 
% We also conduct comprehensive ablation studies to investigate how our method can leverage the predictability of zero-cost metrics to boost the distillation performance.

\paragraph{Main Contributions:}
\begin{itemize}%\vspace{-0.05in} 
    \item We focus on a training-free architecture search for ViTs across multiple datasets. We build ViT-Bench-101 and discover the failures in the generalization of existing training-free methods.

    \item We propose a from-scratch proxy search framework, Auto-Prox, designed to eliminate the need for manual intervention and enhance generalizability.  We propose a joint correlation metric to evolve different zero-cost proxy candidates and an elitism-preserve strategy to improve search efficiency.

    \item We experimentally validate that Auto-Prox achieves state-of-the-art performance across multiple datasets and search spaces, advancing the broader application of ViTs in vision tasks.

\end{itemize}

\section{Related Work}

Vision Transformer (ViT) \cite{vit} has shown remarkable performance in various visual recognition tasks due to its ability to capture long-range dependencies. Recently, researchers have developed several automated techniques to discover more effective ViT architectures. For instance, AutoFormer~\cite{chen2021autoformer} utilizes a one-shot NAS framework for ViT-based architecture search, while BossNAS~\cite{Li2021BossNASEH} incorporates a hybrid CNN-transformer search space along with a self-supervised training scheme. S3~\cite{Chen2021SearchingS3TS} proposes using neural architecture search to automate the design of superior ViT architectures, considering not just the architecture search but also the search space itself. 
ViTAS~\cite{Su2021ViTASVT} has developed a cyclic weight-sharing mechanism for token embeddings of ViTs to stabilize the training of Superformer and prevent catastrophic failures. 
Despite significant progress in ViT architecture search, the aforementioned one-shot-based methods are still computationally demanding. TF-TAS~\cite{DSS} stands as the first method to conduct a training-free architecture search for ViTs, assessing ViT by merging two theoretical perspectives: synaptic diversity from multi-head self-attention layers (MSA) and synaptic saliency from multi-layer perceptrons (MLP). However, current training-free ViT architecture search methods struggle to generalize across various domains. In this paper, we introduce an automated approach to pinpoint excellent proxies for ViTs across diverse datasets. Comparing with EZNAS: 
\ourmethod{} differs notably from EZNAS~\cite{akhauri2022eznas} from the following aspects: (1) EZNAS is only designed for CNNs, but our approach is customized for ViTs. (2) EZNAS simply follows the search strategy from AutoML-Zero. In contrast, we propose an elitism-preserve strategy that significantly improves search efficiency and results (see Figure 5). (3) Our proposed Joint Correlation Metric enhances ranking consistency across multiple datasets, encompassing both small and large-scale datasets, such as ImageNet-1K. In contrast, EZNAS's testing is limited to small datasets.

% %Training-free NAS leverages zero-cost proxies to predict the performance of various neural network architectures without training.
% These zero-cost proxies operate on the principle that certain intrinsic properties or features of an architecture can provide insights into its potential efficacy, even before it is trained.
% Most existing zero-cost proxies are tailored for Convolution-based architectures, such as Synflow~\cite{syflow}, SNIP~\cite{snip}, and GraSP~\cite{grasp}.
% These methodologies rely on gradient computations using a single minibatch of data at initialization and can be computed rapidly.
% Zen-NAS~\cite{ZenNAS} introduced an innovative zero-cost proxy, termed Zen-Score, to evaluate a network's expressivity.
% NWOT~\cite{NWOT} investigates the correlations between linear maps induced by data points for untrained network architectures.
% Despite the promise shown by these proxies, their design often requires expertise and can involve repetitive tasks.
% Our method aims to streamline the discovery process, eliminating the need for expertise and redundant labor in the zero-cost proxy design process.

\section{Methodology}
\label{section:method}
In this section, we first introduce the search space of our automatic zero-cost proxy.
We then delve into the details of the joint correlation metric and the elitism-preserve strategy during the evolutionary process.
Subsequently, we give an analysis of the searched zero-cost proxy.
Finally, we illustrate the training-free ViT search process.

\subsection{Search Space of Automatic Zero-cost Proxy}

To guarantee the effectiveness and flexibility of the zero-cost proxy search space, we begin with revisiting the formulations of existing zero-cost proxies. 
Using this foundational understanding, we design a search space that encompasses eight input candidates, and $56$ primitive operations.
This comprehensive search space allows us to explore a wide range of zero-cost proxies, uncovering potential ones that previous hand-crafted approaches may have overlooked.

\paragraph{Review of Existing Zero-cost Proxies}
\label{sec:review_zc}
To investigate the design of zero-cost proxies, we summarize the existing ones in Table \ref{tab:review_zc}. 
Among these zero-cost proxies, Fisher \cite{Theis2018FasterGP}, SNIP \cite{snip}, Plain \cite{Mozer1988SkeletonizationAT}, and SynFlow \cite{syflow} are conducted on ReLU-Conv2D-BatchNorm2D blocks in CNNs.
In contrast, TF-TAS \cite{DSS} is based on transformer layers in ViTs.
The inputs of these zero-cost proxies derive from the following types of network statistics: Activation (A), Gradient (G), and Weight (W).
Specifically, Fisher computes the sum over all gradients of the activations $\frac{\partial \mathcal{L}}{\partial z}$ in the network, which can be used for channel pruning. 
SNIP, employing weight $\theta$ and gradient $\frac{\partial \mathcal{L}}{\partial \theta}$ as inputs, computes a saliency metric at initialization using a single mini-batch of data.
This metric approximates the change in loss when a specific parameter is removed.
SynFlow, also using weight $\theta$ and gradient $\frac{\partial \mathcal{R}}{\partial \theta}$ as inputs, introduces a modified version of synaptic saliency scores to prevent layer collapse during parameter pruning. 
TF-TAS considers the weights $\theta_l, \theta_k$ and their gradients $\frac{\partial \mathcal{L}}{\partial \theta _l},\frac{\partial \mathcal{L}}{\partial \theta _k}$ from multi-head self-attention (MSA) and multi-layer perceptron (MLP) as inputs, respectively. 
$\left\| \right\| _n$ is the Nuclear-norm.

\begin{table}[t]
\label{tab:zero_cost_proxies}
\centering
\resizebox{1\linewidth}{!}{
\begin{tabular}{c|c|c}
\toprule[1.1pt]
\textbf{Input} & \textbf{Proxy} & \textbf{Formula} \\ \midrule[1.1pt]
% A & Fisher~\cite{Theis2018FasterGP} & $\left\|\partial^2 L/\partial A^2\right\|$ \\ 
A\&G & Fisher~\cite{Theis2018FasterGP} & $\sum_{z\in A}{\left( \frac{\partial \mathcal{L}}{\partial z}z \right) ^2}$ \\ \midrule
W\&G & SNIP~\cite{snip} & $\left| (\frac{\partial \mathcal{L}}{\partial \theta})\odot \theta \right|$ \\  \midrule
W\&G & Plain~\cite{Mozer1988SkeletonizationAT} & $(\frac{\partial \mathcal{L}}{\partial \theta})\odot \theta 
$ \\  \midrule
W\&G & SynFlow~\cite{syflow} & $(\frac{\partial \mathcal{R}}{\partial \theta})\odot \theta, \mathcal{R}=\mathds{1}^T\left(\prod_{\theta_i \in W}\left|\theta_i\right|\right) \mathds{1} $ \\  \midrule
W\&G & TF-TAS~\cite{DSS} & $\sum_{\theta _l}{\left\| (\frac{\partial \mathcal{L}}{\partial \theta _l}) \right\| _n\odot \left\| \theta _l \right\| _n}+\sum_{\theta _k}{(\frac{\partial \mathcal{L}}{\partial \theta _k})\odot \theta _k} $ \\ \bottomrule[1.1pt]
\end{tabular}}
\caption{Overview of existing zero-cost proxies. A, W, and G refer to Activation, Weight, and Gradient.}
\label{tab:review_zc}
    % \vspace{-1em}
\end{table}

\begin{figure}[t]
    \centering
    \includegraphics[width=\linewidth]{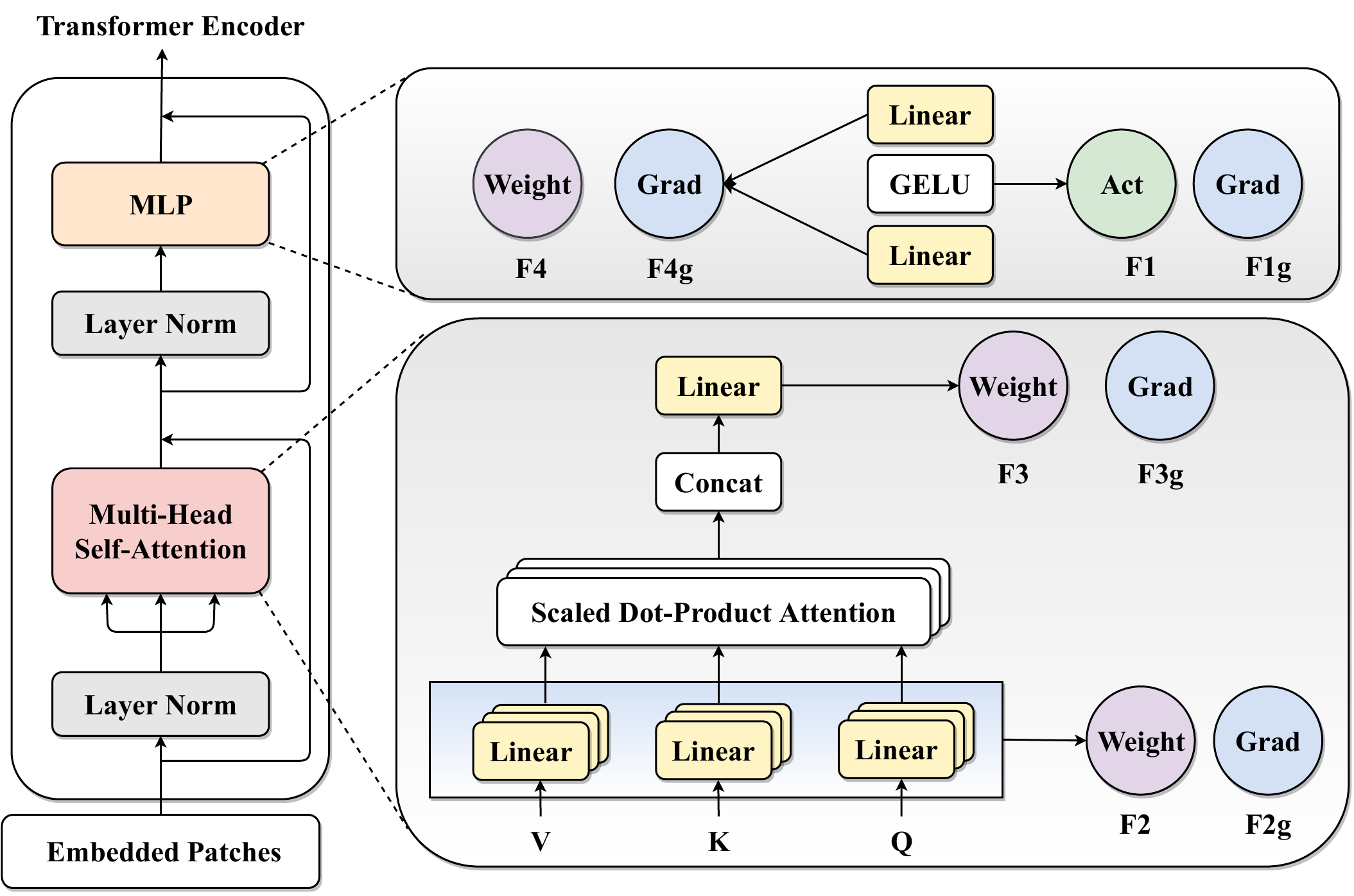}
    % \vspace{-1em}
    \caption{ViT statistics of the zero-cost proxy search space, including activation (Act), gradient (Grad), and weights (Weight) from MSA and MLP modules.}
    \label{fig:vit_statistics}
        % \vspace{-1em}
\end{figure}

\paragraph{ViT Statistics as Input} 
% Based on the stuies in 
The input subsection in Table~\ref{tab:review_zc} explores the input choices for constructing zero-cost proxies, in which activations (A), gradients(G), and weights(W) are identified as the most informative and effective options. 
Activation provides insight into the data distribution and internal network representations, while gradients highlight the sensitivity of weights to the loss functions, and weights reflect the significance of network parameters. 
Building on this observation, we register the activation, weights, and corresponding gradients from each transformer layer as potential inputs of our zero-cost proxy.
For modules in the multi-head self-attention (MSA), we collect their weights and gradients at initialization using a mini-batch of data. 
Regarding multi-layer perception (MLP), we consider weights, gradients, activation, and activation gradients as possible inputs. 
To differentiate between these inputs, we employ symbols like F1 and F1g.
As depicted in Figure~\ref{fig:vit_statistics}, for a transformer layer, there are a total of eight potential inputs in the zero-cost proxy search space.

\paragraph{Primitive Operations} 
To efficiently aggregate information from different types of inputs, we search among different primitive operations to produce the final scalar output.
In the context of the zero-cost proxy, primitive operations are used to process ViT statistics, resulting in the zero-cost proxy score for performance evaluation.
We consider two types of primitive operations, including unary operations (operations with only one operand) and binary operations (operations with two operands).
Inspired by AutoML-based methods~\cite{li2022autoloss, real2020automlzero,dong2023emq}, we provide a total of $24$  unary operations and four binary operations to form the zero-cost proxy search space. 
Since the intermediate variables can be scalar or matrix, the total number of operations is $56$.
All the primitive operations in the zero-cost proxy search space are described in Appendix \textcolor{red}{G}.

\paragraph{Zero-cost Proxy as Computation Graph.} 
% we first revisit inputs and formulations of existing zero-cost proxies, based on which we design a general computation graph that covers different types of zero-cost proxies.
The automatic zero-cost proxy is represented as a computation graph, in which the input nodes are ViT statistics, and the intermediate nodes are primitive operations.
The graph's output yields the proxy score used for ViT ranking. 
There are four main types of computation graphs, namely Linear, Tree, Graph (DAG), and unstructured memory-based structures (such as Automl-zero~\cite{real2020automlzero}). 
The expressiveness of the computation graph increases from Linear to unstructured memory-based structures, but the valid structure in the search space decreases. 
To balance the trade-off between expressiveness and validity, we employ an expression tree to represent the automatic zero-cost proxy.
Based on previous works such as SNIP~\cite{snip}, and TF-TAS~\cite{DSS}, most proxies typically require two types of inputs. 
Therefore, we build an expression tree with two inputs (see Figure \ref{fig:main}). 
The expression tree is applied to every transformer layer of ViT, with the final proxy score derived by averaging the outputs from all transformer layers.

\begin{algorithm}[t]
\small
\caption{Evolutionary Search for Auto-Prox}
\label{alg:evolution}
\textbf{Input}: Search space $\mathcal{S}$, population $\mathcal{P}$, max iteration $\mathcal{T}$, sample ratio $r$, sampled pool $\mathcal{R}$, topk $k$, margin $m$.

\textbf{Output}: Auto-prox with best JCM.

\begin{algorithmic}[1]
\STATE $\mathcal{P}0$ := Initialize population$(P_i)$;
\STATE Sample pool $\mathcal{R}$ := $\emptyset$;
\FOR{$i = 1,2,\ldots,\mathcal{T}$}
\STATE Clear sample pool $\mathcal{R}$ := $\emptyset$;
\STATE Randomly select $\mathcal{R} \in \mathcal{P}$;
\STATE Candidates ${G_i}{k}$ := GetTopk($\mathcal{R}$, $k$);
\STATE Parent $G_i^p$ := RandomSelect$({G_i}_{k})$;
\STATE Mutate $G_i^m$ := MUTATE($G_i^p$);
\STATE // Elitism-Preserve Strategy.
\IF{$\mathrm{JCM}(G_i^m) - \mathrm{JCM}(G_i^{p}) \geq m $}
\STATE Append $G_i^m$ to $P$;
\ELSE
\STATE Go to line 8;
\ENDIF
\STATE Remove the zero-cost proxy with the lowest JCM.
\ENDFOR
\end{algorithmic}
\end{algorithm}

\subsection{Evolving Automatic Proxy on Multiple Datasets}

% In this section, we propose an evolutionary algorithm for automatic proxy search 
Figure \ref{fig:main} presents the automatic proxy search.
At initialization, we randomly sample a population of $N$ candidate zero-cost proxies from the search space, ensuring that they are valid primitively. 
Each of these proxies is then evaluated using the proposed Joint Correlation Metric, which serves as the fitness measure in our evolutionary search process. 
In each iteration of the evolutionary process, the top-k candidates with the highest JCM scores are selected. 
From this subset, a parent is randomly picked for mutation.
When conducting mutation, a random point in the computation graph is chosen and mutated using new inputs or primitive operations.
To guarantee the quality of mutated zero-cost proxies and stave off degradation, we've introduced the Elitism-Preserve Strategy.
This strategy involves judgment and selectively retaining only valid and promising zero-cost proxies for the next generation.
We repeat this process for $T$ iterations to identify the target zero-cost proxy.
In the following, we introduce details of the Joint Correlation Metric and Elitism-Preserve Strategy.

\paragraph{Joint Correlation Metric (JCM)}
Instead of measuring ranking correlation in just one dataset, we propose the Joint Correlation Metric (JCM), which measures the generalization of proxies $\mathcal{Q}$ among multiple datasets. 
We use $M$ datasets $\{\mathcal{D}_i\}_{i=0}^{M}$ and the weight $\{\alpha_i\}_{i=0}^{M}$ to measure the corresponding importance of datasets.
JCM is formulated as the weighted sum of the ranking correlation (Kendall's Tau $\tau$) as follows:
\begin{equation}
    JCM(\mathcal{Q})=\frac{1}{M}\sum_i^M\alpha_i \times \tau(\mathcal{D}_i, \mathcal{Q})
\end{equation}
where $\tau$ is the ranking correlation between the zero-cost proxy score and actual accuracy on dataset $\mathcal{D}_i$.

\begin{figure*}[t]
    \centering
    \includegraphics[width=0.9\linewidth]{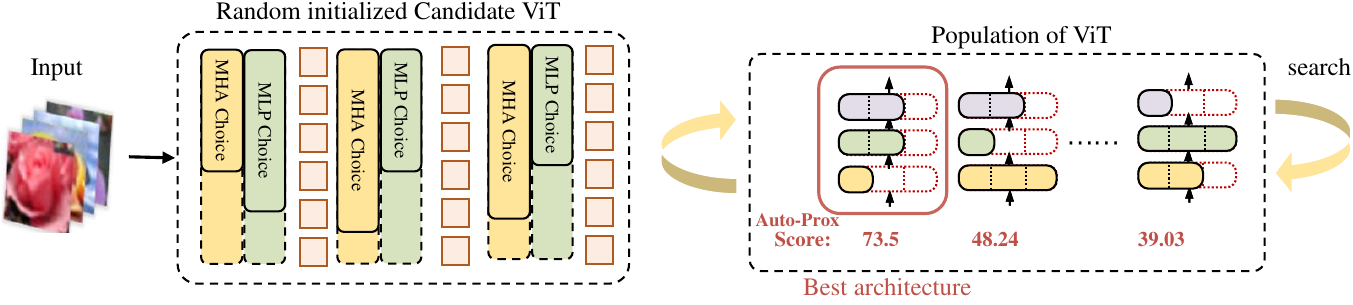}
    \caption{Illustration of the training-free ViT search process. Left: The search space of ViT, where multi-self attention and multi-layer perceptron are mutable. Right: Traverse the search space and select the best ViT with the highest Auto-Prox score.}
    \label{fig:vit}
        % \vspace{-1em}
\end{figure*}

\paragraph{Elitism-Preserve Strategy}
To prevent population deterioration and premature convergence, 
the proposed Elitism-Preserve Strategy involves comparing the performance of the parent and newly generated offspring to measure the validity of the mutation. 
If the offspring is invalid or if its performance is lower than that of the parent, the mutation is considered as deterioration, and therefore new offspring are generated as a replacement. 
Algorithm~\ref{alg:evolution} presents the detailed process.
% Based on Elitism-Preserve Strategy, the search process proceeds to guarantee promising zero-cost proxies.
% This process is repeated to identify the promising proxy for $T$ iterations.
% In evaluating the fitness of a candidate proxy, we adopt the Joint Correlation Metric (JCM). 

% During the process of evolutionary searching, a large number of architectures sampled from the benchmark must be evaluated for each proxy, requiring significant computation. Through our observation of failure proxies, we have identified a common characteristic, namely, a lack of strength in the discrimination ability. 
% Based on this observation, we propose the Discrimination-aware Early Termination (DET) strategy to distinguish between good and bad proxies and to terminate bad ones early to reduce computation. 
% The DET strategy aims to improve the efficiency of the evolutionary algorithm by allowing early identification of proxies with insufficient discrimination ability and eliminating them from further evaluation.

\subsection{Analysis of the Searched Zero-cost Proxy}

Auto-Prox needs ViT statistics as its input. 
To eliminate the potential bias introduced by ViTs from a specific design space, we sampled ViT-accuracy configurations from two distinct design spaces of ViT-Bench-101 and performed zero-cost proxy searches on each.  
As a result, we discovered two separate zero-cost proxies, AutoProxA and AutoProxP, which are used independently to score ViTs sampled from their corresponding search spaces.
It is important to note that Auto-Prox is an automated, from-scratch method, making it versatile across different ViT search spaces and datasets.
Below, we present the formulas for the searched proxies within the AutoFormer (AutoProxA) and PiT search spaces (AutoProxP):
\begin{equation}
    {\rm AutoProxA}=|\frac{\partial \mathcal{L}}{\partial \theta _l}|+\frac{1}{n+\epsilon}\sum_{i=1}^n{sigmoid\left( \frac{\partial \mathcal{L}}{\partial \theta _k} \right)}
\end{equation}
\begin{equation}
  {\rm AutoProxP}=\left\| sigmoid\left( \theta _l \right) \right\| _F-\log \left( \frac{\exp \left( \left\| \theta _k \right\| \right)}{\sum_i{\exp \left( \left\| \theta _k \right\| \right)}} \right) 
\end{equation}
where $\theta _l$ is the weight parameter matrix of QKV layers in the multi-head self-attention (MSA) module, $\frac{\partial \mathcal{L}}{\partial \theta _l}$ is the corresponding gradient matrix.
$\theta _k$ is the weight parameter matrix of linear layers in the MLP module.
$\frac{\partial \mathcal{L}}{\partial \theta _k}$ represents the corresponding gradient matrix.
$n$ is the number of elements.
% in $sigmoid\left( \frac{\partial \mathcal{L}}{\partial \theta _k} \right)$.
$\epsilon$ is the constant set as $1e$-$9$.
$\sum_i$ means the sum of elements in the i-th dimension.
$\left\| \right\| _F$ means the Frobenius-norm.

% $m$ indicates the m-th linear layer in the MSA module.
% we denote the weight parameter matrix of an MSA module as .
% $\mathit{numel}(\theta) = \prod_{i=1}^{n} d_i$ and it denotes the total number of elements in the weight $\theta$, and $d_i$ is the size of the $i$-th dimension. The $\mathcal{R}=\mathbb{1}^T\left(\prod_{\theta_i \in \theta}\left|\theta_i\right|\right) \mathbb{1}$ denotes synaptic flow loss proposed by Synflow~\cite{syflow}.

By analyzing the formulas, we found that high-performing ViTs correlate positively with the following factors: 
(1) A larger norm of weight parameters or gradients in  QKV layers of the MSA module, which approximately indicates the diversity of MSA~\cite{dong2021attention}. 
(2) More salient weight parameters in linear layers of the MLP module, implying a significant impact on performance.
Moreover, we compare the searched proxies with existing TF-TAS ~\cite{DSS}, which is hand-crafted and is included in our zero-cost proxy search space.
Table~\ref{tab:nas} and Table \ref{tab:rank} have demonstrated the superiority of \ourmethod{}, which significantly outperforms the TF-TAS method and meanwhile enjoys more search efficiency, further emphasizing the advantages of automatic searching.
% is requires less time during evaluation.

\subsection{Training-free ViT Search}
\label{sec:evolutionary-framework}
Once we have obtained a good zero-cost proxy through the evolutionary search, we utilize it to perform a training-free ViT search. 
Figure \ref{fig:vit} presents the training-free ViT search process.
Specifically, we randomly explore a large number of candidate architectures from the ViT search space.
% (AutoFormer~\cite{chen2021autoformer} or PiT~\cite{DSS}).
Then, we evaluate each ViT architecture and select the best one based on the searched zero-cost proxy (Auto-Prox) score.
Without the need for costly and time-consuming training, the ViT search process is highly efficient.

\begin{table*}[t]
% \small
	\centering
	\setlength{\tabcolsep}{3.3pt}
	% \footnotesize
	 \resizebox{1\linewidth}{!}{
	\begin{tabular}{c|c|crr|crr|crr}
		\toprule
		\multirow{2}{*}{Search Space}           & \multirow{2}{*}{Proxy} & \multicolumn{3}{c|}{CIFAR-100} & \multicolumn{3}{c|}{Flowers} & \multicolumn{3}{c}{Chaoyang}                                                                \\
		%\midrule
		\cline{3-11}
&                        & Kendall                     & Spearman                       & Pearson                        & Kendall & Spearman & Pearson & Kendall & Spearman & Pearson \\

		\midrule
		\multirow{6}{*}{AutoFormer}        
  & GraSP      &  0.84$_{\pm 0.73}$     & 1.35 $_{\pm 0.92}$                          &  0.82 $_{\pm 1.58}$      & 0.82$_{\pm 1.58}$   & -7.33$_{\pm 0.10}$     & -4.14$_{\pm 0.84}$     & -4.42$_{\pm 0.38}$    & -6.53$_{\pm 0.36}$      &  6.26$_{\pm 0.72}$    \\
& SynFlow     & 37.66$_{\pm 0.63}$    &  52.89$_{\pm 1.11}$                          &   52.01$_{\pm 0.74}$    & 62.59$_{\pm 0.01}$   & 82.13$_{\pm 0.01}$    & 62.26$_{\pm 7.71}$     & 27.87$_{\pm 0.75}$    & 39.30$_{\pm 1.51}$      &  41.08$_{\pm 1.14}$   \\
& TENAS      & -30.03$_{\pm 0.28}$      & -43.27$_{\pm 0.46}$                           & -42.66$_{\pm 0.26}$                            & -53.55$_{\pm 0.05}$   & -73.79$_{\pm 0.11}$    & -54.17$_{\pm 8.31}$    & -27.81$_{\pm 0.11}$    &  -39.69$_{\pm 0.24}$     & -40.79$_{\pm 0.18}$    \\
& NWOT                   &  54.65$_{\pm 0.22}$     & 63.11$_{\pm 0.26}$                          &    60.01$_{\pm 0.17}$                        & 68.16$_{\pm 0.03}$   & 82.06$_{\pm 1.07}$     & 53.91$_{\pm 7.46}$   & 27.29$_{\pm 0.25}$   &  38.96$_{\pm 0.53}$   &  40.57$_{\pm 0.56}$   \\
& TF-TAS                   &  35.89$_{\pm 0.26}$        & 50.84$_{\pm 0.58}$                           &  50.51$_{\pm 0.46}$                             & 63.90$_{\pm 0.04}$   & 83.28$_{\pm 0.09}$    & 62.80$_{\pm 8.58}$    &  27.14$_{\pm 0.35}$   & 38.52$_{\pm 0.68}$  
& 40.57$_{\pm 0.40}$     \\
& Ours                   & \textbf{55.67}$_{\pm 0.74}$   &    \textbf{63.87}$_{\pm 1.04}$                       &   \textbf{60.56}$_{\pm 0.93}$                        & \textbf{69.19}$_{\pm 2.03}$  & \textbf{83.65}$_{\pm 0.93}$     & \textbf{79.52}$_{\pm 7.15}$     & \textbf{33.76}$_{\pm 0.46}$    & \textbf{41.76}$_{\pm 0.66}$     &  \textbf{42.63}$_{\pm 0.45}$    \\
		\midrule
		\multirow{6}{*} {PiT}     
  & GraSP                    &  -42.02$_{\pm 0.58}$                       &  -58.71$_{\pm 0.74}$                         &  -31.53$_{\pm 0.09}$     & -50.66$_{\pm 0.07}$   & -69.64$_{\pm 0.09}$    & -40.90 $_{\pm 0.57}$    & -16.00$_{\pm 0.61}$    & -22.94$_{\pm 1.33}$    & -19.07$_{\pm 0.56}$     \\
& SynFlow                    &  69.79$_{\pm 1.16}$                     &    87.05$_{\pm 0.77}$                       &  70.80$_{\pm 0.08}$       & 62.22$_{\pm 2.38}$   & 79.98$_{\pm 2.16}$    & 71.66$_{\pm 0.99}$     &  30.96$_{\pm 4.38}$  &  42.66$_{\pm 8.46}$     &  39.24$_{\pm 5.89}$    \\
& TENAS                  &  -2.13$_{\pm 0.30}$     & -3.21$_{\pm 0.74}$                        &  -1.68$_{\pm 0.17}$                           & -2.86$_{\pm 0.63}$  & -4.23$_{\pm 1.46}$   & -3.33$_{\pm 1.49}$    &  -3.34$_{\pm 0.03}$  & -5.04$_{\pm 0.07}$      & -3.55$_{\pm 0.13}$    \\
& NWOT                   & -2.61$_{\pm 0.01}$      & -4.13$_{\pm 0.01}$                         &   -0.52$_{\pm 1.04}$                         & 2.67$_{\pm 0.23}$  & 3.69$_{\pm 0.49}$   & 0.71$_{\pm 0.46}$   & 4.73$_{\pm 0.16}$   & 6.87$_{\pm 0.31}$   & 4.43$_{\pm 0.26}$     \\
& TF-TAS                   & 63.83$_{\pm 0.06}$     & 82.20$_{\pm 0.04}$                          & 58.37$_{\pm 1.84}$                          & 64.48$_{\pm 0.08}$  & 82.91$_{\pm 0.06}$    &  67.23$_{\pm 0.79}$   &  37.99$_{\pm 1.54}$  & 52.92$_{\pm 2.54}$   & 42.68$_{\pm 0.62}$    \\
& Ours                   &  \textbf{82.07}$_{\pm 0.33}$    &  \textbf{95.12}$_{\pm 0.09}$    & \textbf{73.25}$_{\pm 1.30}$                           & \textbf{79.25}$_{\pm 0.94}$    & \textbf{92.94}$_{\pm 0.33}$    & \textbf{79.53}$_{\pm 0.14}$    & \textbf{41.67}$_{\pm 4.45}$    & \textbf{55.09}$_{\pm 7.50}$    & \textbf{49.17}$_{\pm 3.94}$     \\
		\bottomrule
	\end{tabular}
	}
	%\vspace{-0.2cm}
 \caption{Ranking correlation results (\%) on CIFAR-100, Flowers, and Chaoyang. Auto-Prox achieves the highest ranking correlation with distillation accuracy on all three datasets of the ViT-Bench-101, demonstrating superior generalization.}
	% \label{tab:nb201_sota}
 \label{tab:rank}
\end{table*}

\begin{table*}[t]
% \small
% \vspace{2mm}
 \centering
	\setlength{\tabcolsep}{3.3pt}
	% \footnotesize
	 \resizebox{1\linewidth}{!}{
    \begin{tabular}{c|c|crr|crr|crr}
    \toprule
    \multicolumn{1}{l|}{\multirow{2}[4]{*}{Search Space}} & \multirow{2}[4]{*}{Proxy} & \multicolumn{3}{c|}{CIFAR-100} & \multicolumn{3}{c|}{Flowers} & \multicolumn{3}{c}{Chaoyang} \\
\cmidrule{3-11}          &       & \multicolumn{1}{l}{Param(M)} & \multicolumn{1}{l}{Dis.Acc(\%)} & \multicolumn{1}{l|}{Search Cost} & \multicolumn{1}{l}{Param(M)} & \multicolumn{1}{l}{Dis.Acc(\%)} & \multicolumn{1}{l|}{Search Cost} & \multicolumn{1}{l}{Param(M)} & \multicolumn{1}{l}{Dis.Acc(\%)} & \multicolumn{1}{l}{Search Cost} \\
    \midrule
    \multirow{7}[2]{*}{AutoFormer} & Random & 8.30      & 76.72      & N/A       & 6.18      & 67.64      &  N/A      & 6.54      & 84.20      & N/A  \\
          & GraSP &  5.77    &  77.53     &  2.62 h     & 5.29      & 66.08      & 2.71 h      & 6.12      & 84.81      & 1.52 h \\
          & SynFlow &  9.52     &  77.83     & 1.91 h      &  8.13     & 68.61      & 1.85 h     & 5.82      &  83.87     & 1.77 h\\
          & TENAS &   5.40    & 75.54      & 4.83 h     &  5.25     & 66.69      & 4.82 h     & 5.16      &   84.53      & 4.80 h\\
          & NWOT  &  8.36     & 76.79      & 3.24 h     & 8.87      & 69.18      & 3.24 h     & 5.53      & 84.71      & 3.07 h\\
          & TF-TAS   & 5.25      & 75.72      & 1.94 h     & 5.80      & 67.57      & 1.87 h     & 5.60       & 84.57       & 1.92 h\\
          & Ours  & 9.11        &  \textbf{78.26}     & \textbf{0.70} h      & 9.80      & \textbf{69.71}     & \textbf{0.85} h       & 8.97      &  \textbf{84.85}     & \textbf{0.85} h \\
    \midrule
    \multirow{7}[2]{*}{PiT} & Random & 5.33      & 75.84      & N/A     &  4.88     & 65.30      & N/A      & 5.24      &  82.94     & N/A \\
          & GraSP & 4.53      & 76.03      & 1.24 h      & 3.72      &  66.58      & 1.85 h      & 4.63      & 83.87      & 0.86 h \\
          & SynFlow &  11.05     &  77.13     & 1.08 h      &  5.23    &  68.12   &  0.99 h     & 4.93      & 83.73      & 0.70 h\\
          & TENAS &  6.93     &  76.09     &  5.14 h     &  4.26       & 68.03       & 5.14 h      & 6.76      & 83.64      & 5.07 h \\
          & NWOT  &  5.21     & 76.64      &  3.02 h    &   10.77      &  67.72     & 3.09 h      & 6.37      & 83.31      & 3.08 h \\
          & TF-TAS   & 16.07      & 77.06      & 1.21 h      &  10.30   &  68.21    & 0.95 h      &   4.32    &   84.34    & 0.71 h \\
          & Ours  &  6.22     & \textbf{77.26}      &  \textbf{0.51} h     &   6.20      &  \textbf{68.85}   & \textbf{0.39} h     &  4.49     &  \textbf{84.53}      & \textbf{0.31} h  \\
    \bottomrule
    \end{tabular}%
	}
 	\caption{Comparing the distillation performance on three datasets of ViTs sampled from Autoformer and PiT search spaces, Auto-Prox achieves competitive results with the lowest search cost (measured on a single NVIDIA A40 GPU).}
  	\label{tab:nas}
	%\vspace{-0.2cm}
\end{table*}

\section{Experiments}

\subsection{ViT-Bench-101}

ViT-Bench-101 provides ground-truth accuracy of ViTs on both tiny datasets and large-scale datasets.
For the tiny datasets, we employ CIFAR-100 \cite{cifar100}, Flowers \cite{flowers}, and Chaoyang \cite{zhu2021hard}, while for the large-scale datasets, we focus on ImageNet-1K.
Motivated by findings from \cite{li2022locality}, which show that ViTs achieve significant gains on tiny datasets when distilled from an efficient CNN teacher network, we include distillation accuracy for ViTs on these datasets using a given teacher.
Specifically, for the smaller datasets excluding ImageNet-1K, ViT-Bench-101 offers both distillation accuracy and vanilla accuracy for ViTs sampled from AutoFormer and PiT search spaces.
This supports the evaluation of zero-cost proxies based on the score-accuracy correlation in different scenarios, with or without distillation.
Regarding the ImageNet-1K dataset, we follow the demonstration in \cite{chen2021autoformer, DSS}, showing that the sub-nets with inherited weights from the pre-trained AutoFormer supernet can achieve performance comparable to the same network when retrained from scratch. 
Thus, we sample ViTs from the AutoFormer search space and collect their performance by inheriting weights from the publicly available supernet.
ViT-Bench-101 contains 500 ground-truths for each dataset, facilitating fair comparisons between different zero-cost proxies and identifying high-performing ViTs.
For more information on the training details for each dataset included in ViT-Bench-101, please refer to Appendix \textcolor{red}{A}.

\subsection{Implementation Details}

\noindent \textbf{Evolutionary Zero-cost Proxy Search} 
We partition the whole ViT-Bench-101 dataset into a validation set (60\%) for proxy searching and a test set (40\%) for proxy evaluation.
There is no overlap between these two sets. 
% During the proxy search, we randomly select 100 ViTs from the validation set and obtain their ground truths on each dataset. 
% In the evaluation phase, we compare the searched proxy against
% other existing proxies. We randomly sample 100 ViTs from
% the test set to calculate ranking correlations on target datasets.
In the evolutionary search process, we employ a population size of $\mathcal{P}=20$, and the total number of iterations $\mathcal{T}$ is set to $200$. 
To evaluate zero-cost proxy candidates, we randomly sample $100$ ViT ground-truth configurations from ViT-Bench-101 and measure the ranking consistency between its zero-cost proxy score and actual accuracy.
We then calculate the Joint Correlation Metric based on the ranking consistencies on multiple datasets as the fitness function. 
When conducting mutation, the probability of mutation for a single node in a zero-cost proxy representation is set to $0.5$. 
The margin $m$ in the Elitism-Preserve Strategy is $0.1$.
The zero-cost proxy search process is conducted on a single NVIDIA A40 GPU and occupies the memory of only one ViT.

\noindent \textbf{Training-free ViT Search} 
Based on the \ourmethod{} score, the ViT search process is efficient since gradient back-propagation is not included.
we randomly sample 400 ViTs from AutoFormer \cite{chen2021autoformer} and PiT \cite{DSS} search spaces. 
The parameter intervals of ViTs from AutoFormer and PiT search spaces in our experiments are $4\sim9$ M and $2\sim25$ M, respectively. 
The final accuracy of the ViTs with the highest zero-cost proxy scores is reported as the results of the ViT search process. 
% The hyperparameters for ViT retraining are consistent with those used in building ViT-Bench-101.
The hyperparameters for ViT retraining and building ViT-Bench-101 are adopted from \cite{li2022locality}.

\begin{table}[t]
\centering
% \newcommand{\tabincell}[2]{\begin{tabular}{@{}#1@{}}#2\end{tabular}}
% \vspace{-2mm}
\setlength{\tabcolsep}{2.4mm}
\small{
 \resizebox{1\linewidth}{!}{
\begin{tabular}{l|cc|c}
\toprule
% \hline 
Models       & \multicolumn{1}{c}{\tabincell{c}{Param (M)}} & \tabincell{c}{Acc (\%)}  & GPU Days \\
\midrule
% MobileNet-V2~\cite{39} &  3.5                      & -                          & 72.0                       & -                        &    CNN     & Manual           &      -       \\
Deit-Ti~\cite{3}     &  5.7                           &   72.2                       &       -      \\
TNT-Ti~\cite{TNT}     &  6.1                           &  73.9     &       -      \\
ViT-Ti~\cite{2}     &  5.7                           & 74.5                                 &      -       \\
PVT-Tiny~\cite{33} &  13.2                           &  75.1                            &      -       \\
ViTAS-C~\cite{8}    &  5.6                           &  74.7                       & 32 \\
AutoFormer-Ti~\cite{7}&  5.7                           & 74.7                     & 24       \\
TF-TAS-Ti \cite{DSS}    & 5.9                           &  75.3                        &  0.5          \\
\ourmethod{} (Ours)    &  6.4                          & \textbf{75.6}                         &  \textbf{0.1}           \\
\bottomrule
\end{tabular}
}}
\caption{ImageNet results on the AutoFormer search space. 
% $*$ denotes the results reported by \cite{17}.
}
\label{tab:ImageNet}
% \vspace{-2mm}
\end{table}

\begin{table}[htbp]
  \centering
  % \small
  % \resizebox{1\linewidth}{!}{
    \begin{tabular}{l|ccc}
    \toprule
    Method & Kendall    &Spearman  & Pearson    \\
    \midrule
    % Grasp & -2.37$_{\pm 0.05}$ & -7.53$_{\pm 1.30}$ & -2.56$_{\pm 0.08}$ \\
    Snip & 14.61$_{\pm 1.55}$ & 30.62$_{\pm 6.03}$ & 49.45$_{\pm 10.68}$ \\
    SynFlow & 14.81$_{\pm 2.33}$  & 27.69$_{\pm 7.25}$ & 44.19$_{\pm 10.35}$   \\
    NWOT  & 13.34$_{\pm 0.08}$  & 19.79$_{\pm 1.53}$  & 38.39$_{\pm 9.94}$ \\
    TF-TAS  & 14.52$_{\pm 1.74}$ & 29.93$_{\pm 6.38}$ & 48.70$_{\pm 11.04}$ \\
    \ourmethod{} (ours) & \textbf{25.44}$_{\pm 0.90}$ & \textbf{37.10}$_{\pm 1.86}$ & \textbf{51.84}$_{\pm 10.07}$ \\
    \bottomrule
    \end{tabular}%
    % }
      \caption{Ranking Correlation Results (\%) for ImageNet-1K dataset in ViT-Bench-101.}
  \label{tab:zc_proxies}%
\end{table}%

\subsection{Experimental Results on Tiny Datasets}

Table \ref{tab:nas} presents a comparison of distillation results obtained by using various zero-cost proxies on AutoFormer and PiT search spaces. 
In addition, we evaluate the ranking correlations of these zero-cost proxies under different experimental settings using metrics such as Kendall's tau \cite{abdi2007kendall}, Spearman's rho \cite{stephanou2021sequential}, and Pearson's correlation coefficient \cite{bowley1928standard}. 
% To further illustrate the effectiveness of \ourmethod{}, we present the rank consistency results on three datasets in Table~\ref{tab:nb201_sota}. 
As shown in both Table~\ref{tab:rank} and Figure \ref{fig:proxy_score}, \ourmethod{} outperforms other excellent zero-cost proxies.
% demonstrating its effectiveness and superiority in achieving consistent and accurate ViT architecture search. 
These findings demonstrate the importance of using effective zero-cost proxies and the superiority of \ourmethod{} in achieving higher performance in ViT architecture search.

\subsection{Experimental Results on ImageNet-1K}

To validate the effectiveness and superiority of our proposed \ourmethod{} further, we evaluate its performance on the challenging ImageNet-1K dataset, comparing its ranking correlation and top-1 classification accuracy with other hand-crafted and automatically searched ViT methods. 
The results, as presented in Table \ref{tab:ImageNet}, demonstrate that the optimal ViT architecture searched by \ourmethod{} outperforms both excellent hand-crafted and other automatically searched ViT methods, underscoring the superiority of our proposed approach. 
Importantly, our approach strikes a good balance between performance and search efficiency, requiring only 0.1 GPU days for ViT architecture search, making it a practical and efficient approach for ViT architecture search. 
Furthermore, the ranking correlation results in Table \ref{tab:zc_proxies} show that our proposed \ourmethod{} performs better than other zero-cost proxies on the challenging ImageNet task, further validating the effectiveness and generalization of our approach.

\begin{figure}[t]
	\setlength{\abovecaptionskip}{0.cm}
	\setlength{\belowcaptionskip}{-0.cm}
	\centering
	\begin{minipage}[t]{0.49\linewidth}
		\centering
		\includegraphics[width=1\linewidth]{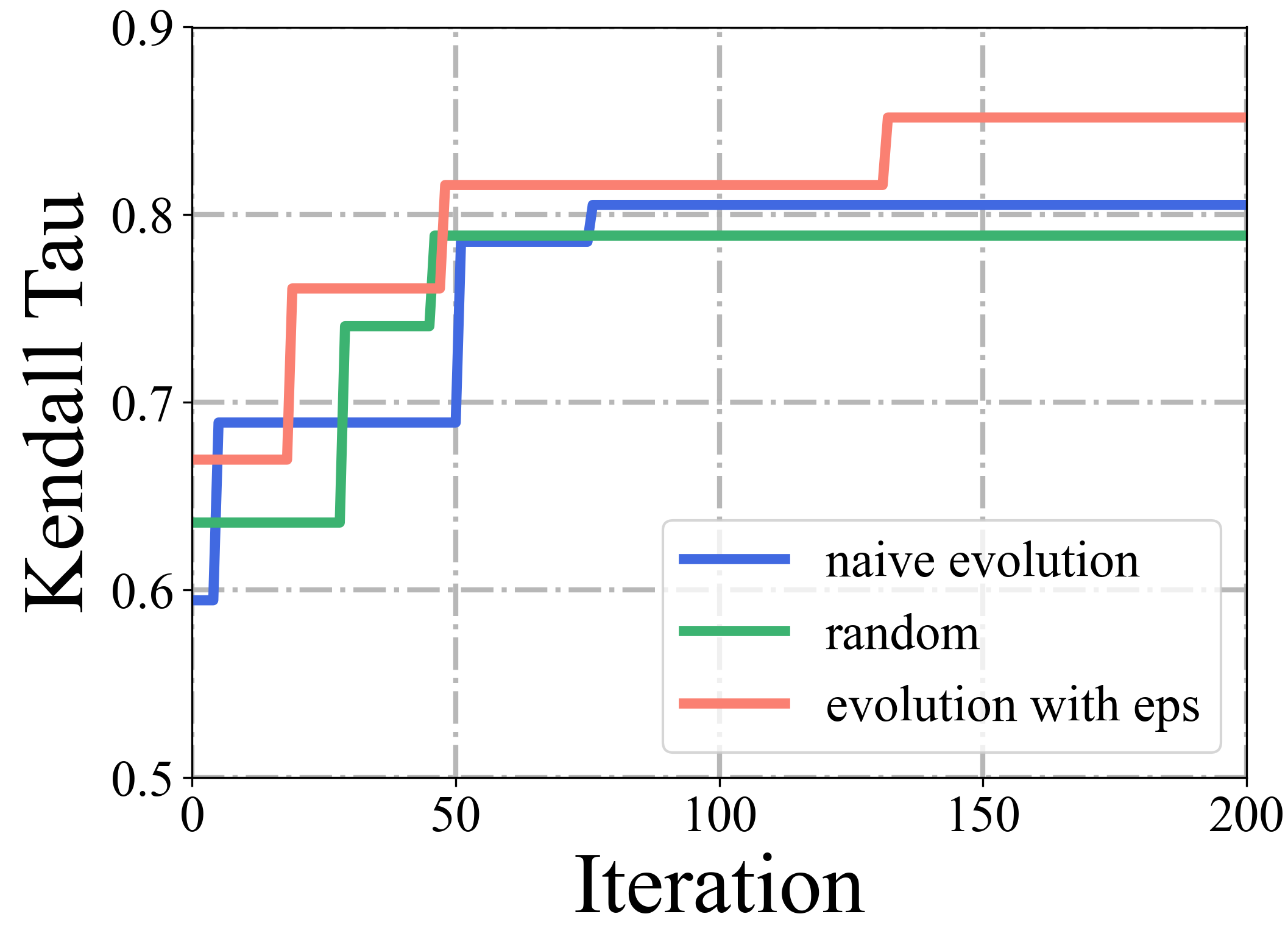}
	\end{minipage}
	\begin{minipage}[t]{0.49\linewidth}
		\centering
		\centering
		\includegraphics[width=1\linewidth]{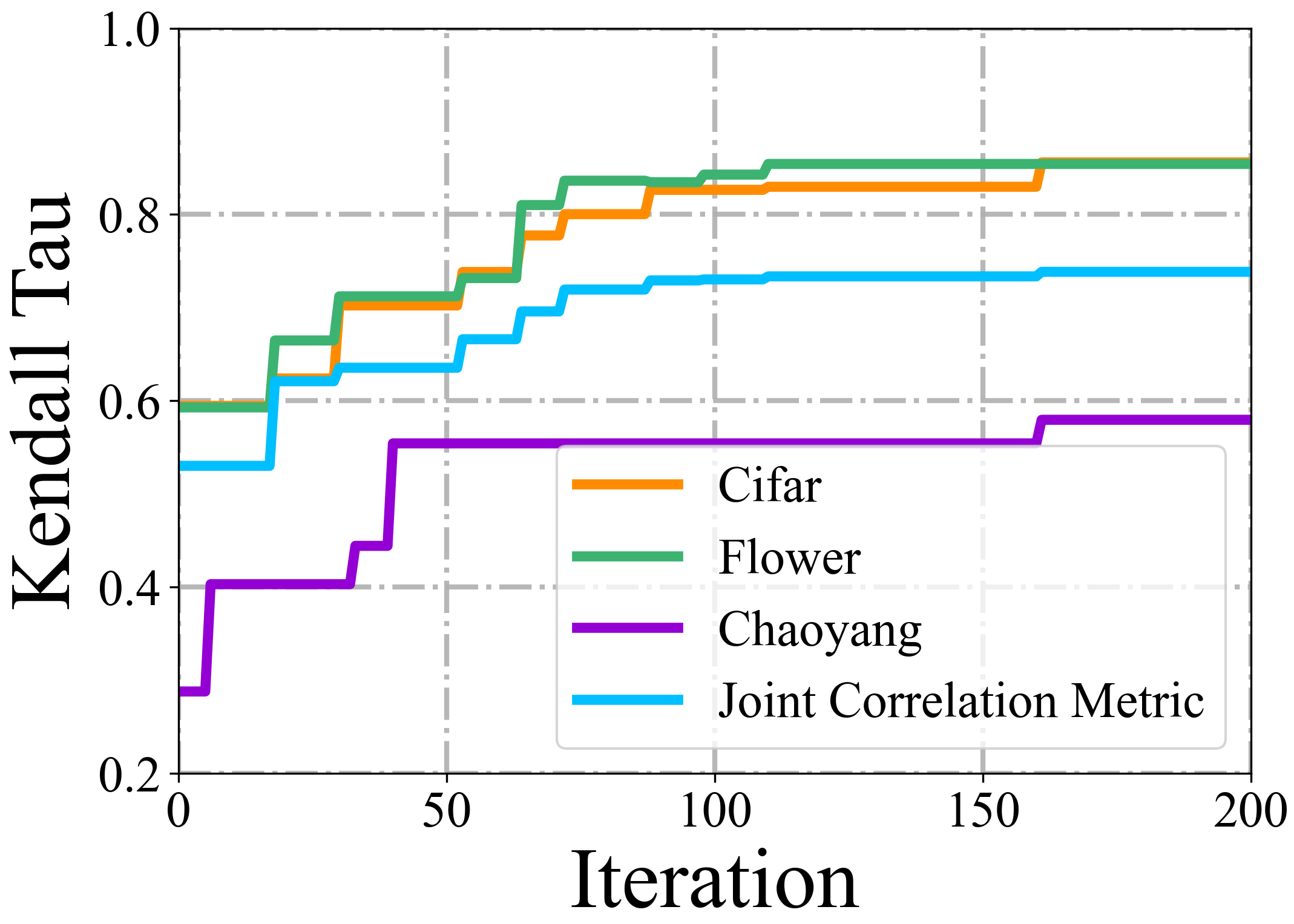}
	\end{minipage}
	\caption{Left: Comparison of naive evolutionary search, random search, and evolutionary search with Elitism-Preserve Strategy, which is denoted as 'eps'.
Right: Evolutionary process of ranking correlations on different datasets, and the proposed joint correlation metric.}\label{fig:ablation}
 % \vspace{-1em}
\end{figure}

\begin{figure}[t]
\setlength{\abovecaptionskip}{0.cm}
\setlength{\belowcaptionskip}{-0.cm}
\centering
\begin{minipage}[t]{0.50\linewidth}
    \centering
    \includegraphics[width=1\linewidth]{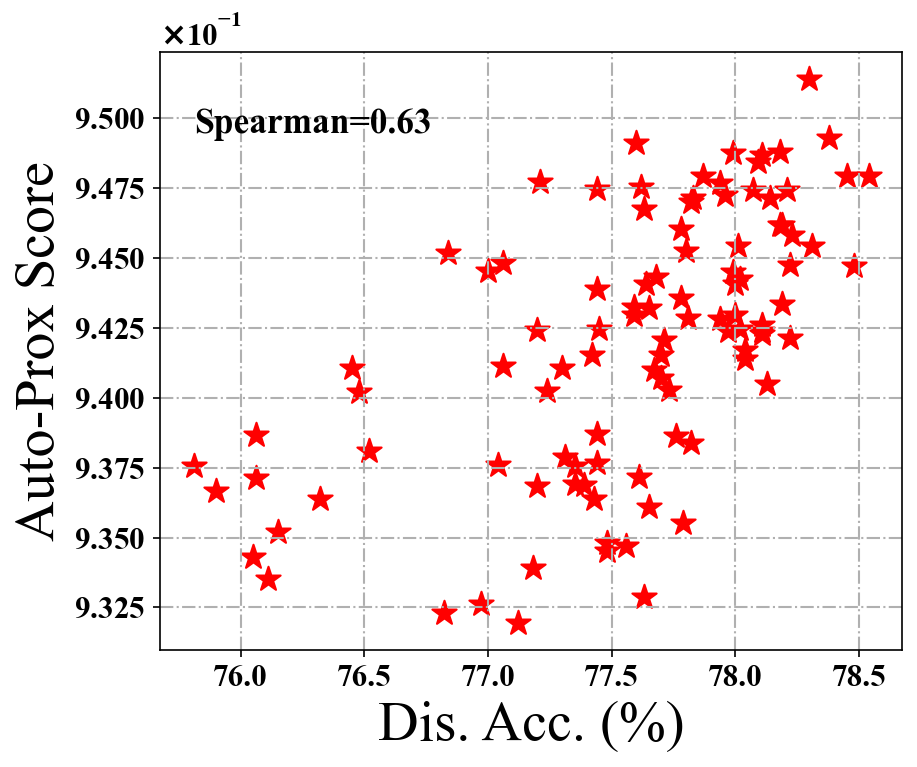}
    % \caption{(a)}
\end{minipage}%
\begin{minipage}[t]{0.50\linewidth}
    \centering
    \includegraphics[width=1\linewidth]{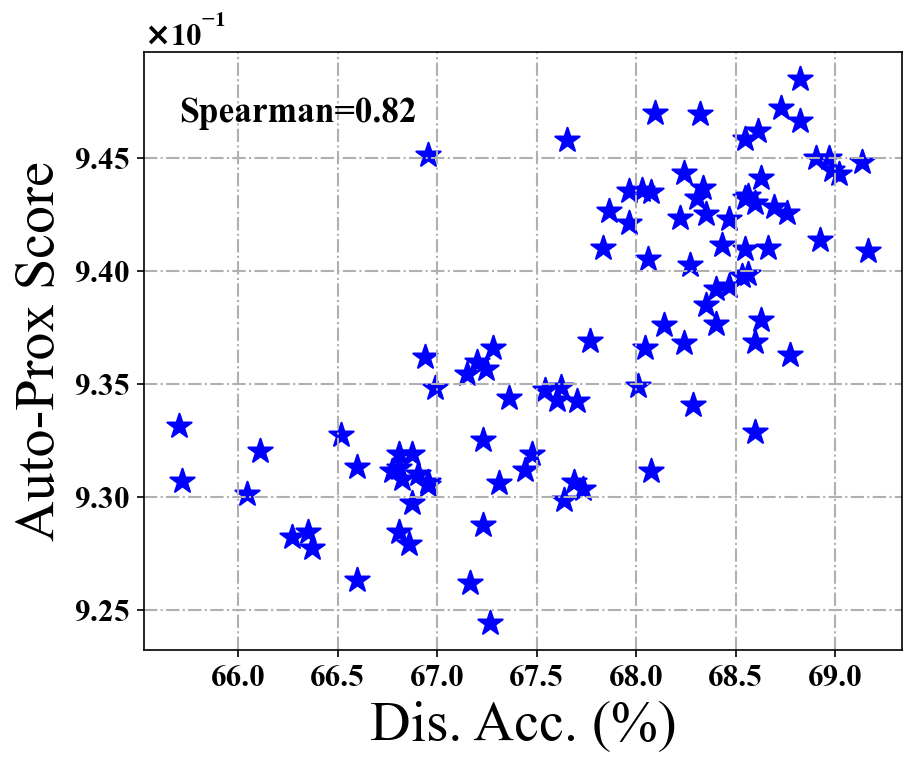}
    % \caption{(b)}
\end{minipage}
\\ 
\begin{minipage}[t]{0.50\linewidth}
    \centering
    \includegraphics[width=1\linewidth]{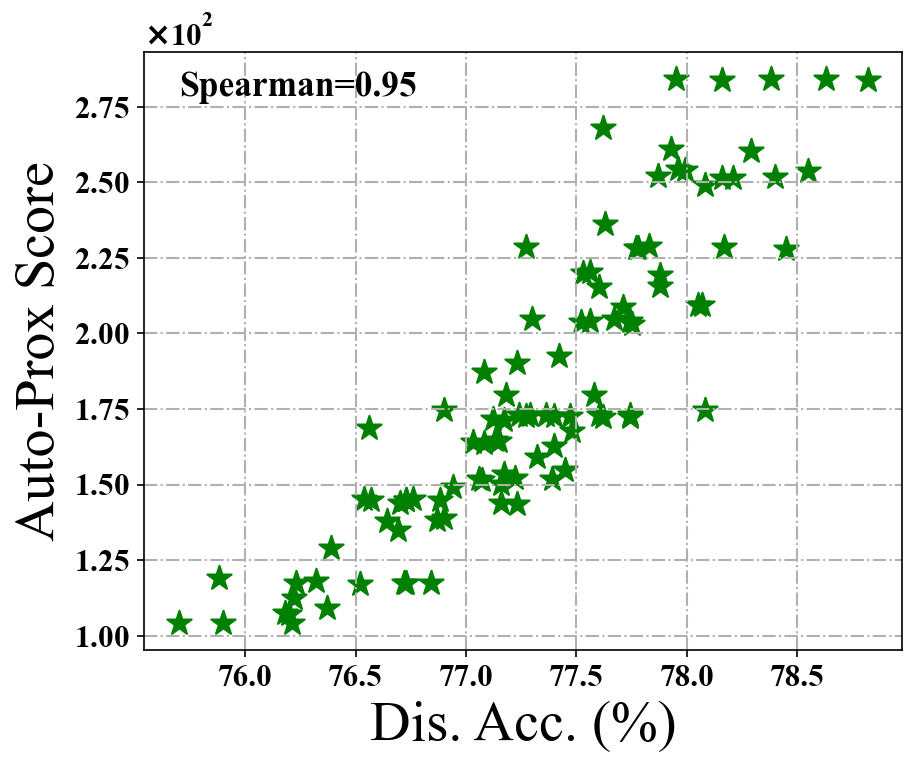}
    % \caption{(c)}
\end{minipage}%
\begin{minipage}[t]{0.50\linewidth}
    \centering
    \includegraphics[width=1\linewidth]{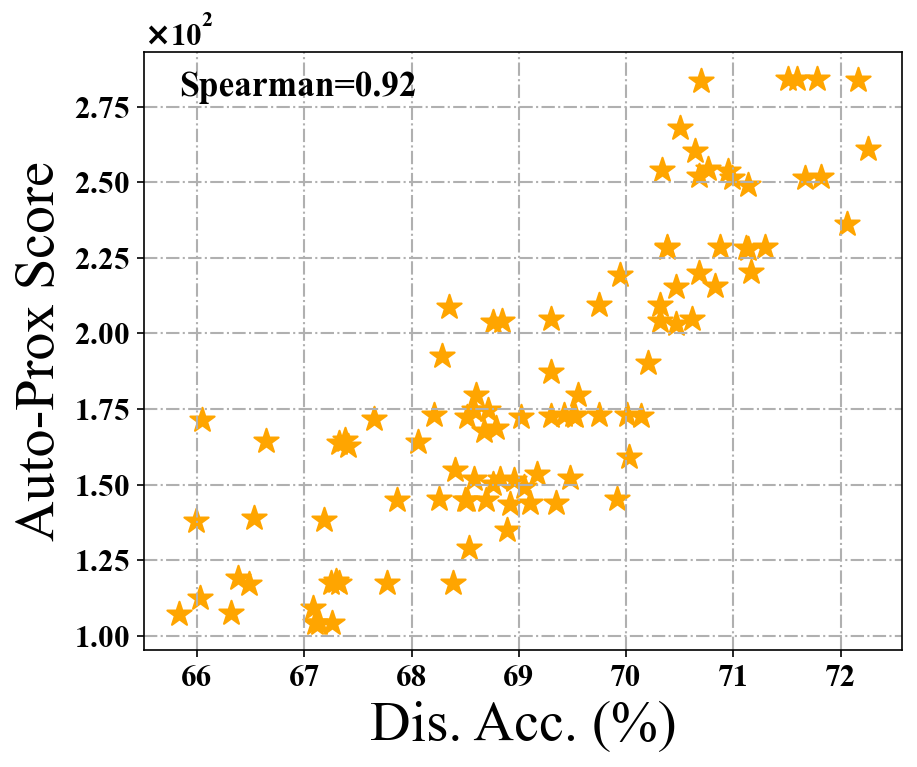}
    % \caption{(d)}
\end{minipage}
 % \vspace{1mm}
\caption{Correlation of distillation accuracy and \ourmethod{} scores on AutoFormer search space (top left: CIFAR-100, top right: Flowers) and PiT search space (bottom left: CIFAR-100, bottom right: Flowers).}\label{fig:proxy_score}
\end{figure}

\begin{table}[t]
  \centering
  % \small
  % \resizebox{1\linewidth}{!}{
    \begin{tabular}{l|ccc}
    \hline
    Method & CIFAR-100    & Flowers  & Chaoyang    \\
    \hline
    % Grasp & -2.37$_{\pm 0.05}$ & -7.53$_{\pm 1.30}$ & -2.56$_{\pm 0.08}$ \\
    SynFlow & 86.88$_{\pm 1.55}$  & 86.57$_{\pm 2.38}$ &  64.51$_{\pm 4.96}$  \\
    NWOT  & -0.45$_{\pm 8.08}$  & 1.98$_{\pm 6.55}$  & -2.26$_{\pm 10.63}$ \\
    TF-TAS  & 84.79$_{\pm 4.57}$ & 86.51$_{\pm 3.74}$  & 69.09$_{\pm 7.05}$ \\
    Ours & \textbf{91.01}$_{\pm 2.63}$ & \textbf{90.70}$_{\pm 1.96}$ & \textbf{76.95}$_{\pm 6.54}$ \\
    \hline
    \end{tabular}%
    % \vspace{-4mm}
      \caption{Spearman correlation results (\%) with vanilla accuracy (without distillation).
    ViT-accuracy configurations are sampled from the PiT search space of ViT-Bench-101.}
  \label{tab:zc_vanilla}%
   % \vspace{-2mm}
\end{table}%

\subsection{Ablation Study}
\label{section:ablation}

As shown in  Figure \ref{fig:ablation} (left), we observe that
the proposed elitism-preserve strategy significantly enhances search efficiency, leading to better results. 
The findings underscore the importance of preserving the top-performing zero-cost proxies during the evolutionary process to achieve better search efficiency. 
Moreover, Figure \ref{fig:ablation} (right) reveals that optimizing the Joint Correlation Metric facilitates the discovery of high-performing zero-cost proxies across multiple datasets.
We also conducted a zero-cost proxy search using vanilla accuracy provided by ViT-Bench-101.
The results presented in Table \ref{tab:zc_vanilla} demonstrate the superior performance of \ourmethod{} in the scenario without distillation.
% The results provide valuable insights into improving the efficiency of architecture search algorithms, especially in the context of ViT architecture search, where the architecture space is vast and complex.

\section{Conclusion}
In this paper, we present Auto-Prox, an automated proxy discovery framework designed for generality across multiple data domains. 
To facilitate the evaluation of zero-cost proxies, we propose the ViT-Bench-101 dataset as a standardized benchmark.
We design a search space of automatic zero-cost proxies for ViTs and develop a joint correlation metric to optimize genetic proxy candidates. 
Additionally, we introduce an elitism preservation strategy to enhance search efficiency.
With the searched proxy, we conduct a ViT architecture search in a training-free manner, achieving significant accuracy gains.  
Extensive experiments validate the efficiency and effectiveness of Auto-Prox across multiple datasets and search spaces.  
We hope this elegant and practical approach will inspire more investigation into the generalization of training-free ViT search methods.

% Auto-Prox searches for more effective zero-cost proxies for ViT across multiple data domains to enhance generalization. 
% We propose the ViT-Bench-101 dataset as a benchmark to evaluate zero-cost proxies.

\section{Acknowledgements}
% This work is supported by the National Natural Science Foundation of China (No.62025208) and the Open Project of Xiangjiang Laboratory (No.22XJ01012).

This work is supported by the National Natural Science Foundation of China (No.62025208), the Open Project of Key Laboratory (2022-KJWPDL-06), and the Xiangjiang Laboratory Fund (No.22XJ01012).

% \bigskip

\appendix
\part*{Appendix}
In this appendix, we provide additional details about \ourmethod{}.
Namely, we first introduce ViT-Bench-101, the dataset used for evaluating different zero-cost proxies.
Then, we elaborate on primitive operations, ranking correlation metrics, computation graph details, and the validity check. 
Additionally, we present more ablation studies, including the sensitivity analysis of seeds, more evolutionary iterations, and application to larger ViTs.
Lastly, we present ViTs that were searched in two search spaces with \ourmethod{}.

\section{Details of ViT-Bench-101}

ViT-Bench-101 evaluates each ViT architecture on four datasets.
It serves as a benchmark for zero-cost proxies by analyzing the ranking correlation between their proxy scores and accuracies, and also enables different NAS algorithms to identify high-performing ViTs.

\subsection{Datasets}

We choose the most popular image classification datasets, including CIFAR-100~\cite{cifar}, Oxford Flowers~\cite{krizhevsky2009learning}, Chaoyang~\cite{zhu2021hard}, and ImageNet-1K~\cite{imagenet}.
CIFAR-100 contains 50,000 training samples and 10,000 test samples across 100 classes.
Oxford Flowers comprises 2,040 training samples and 6,149 test samples for 102 classes.
Chaoyang, which falls within the medical image domain, includes 4,021 training samples and 2,139 test samples, distributed among four classes.
The ImageNet-1K dataset spans 1,000 object classes and consists of 1,281,167 training images, 50,000 validation images, and 100,000 test images.

\subsection{ViT Search Space}
We conduct experiments on AutoFormer \cite{chen2021autoformer}, and PiT \cite{DSS} search spaces. 
Their detailed, searchable components are presented in Table~\ref{tab:Search Space}.

\begin{itemize}
    \item \textbf{AutoFormer:} The variable factors of basic transformer blocks include the embedding dimension (ranging from $192$ to $240$, with a step of $24$),  Q-K-V dimension (ranging from $192$ to $256$, with a step of $64$), number of heads (ranging from $3.5$ to $4$, with a step of $0.5$), and MLP ratio (ranging from $3$ to $4$, with a step of $1$). The network depth varies between 12 and 14 layers.
    \item \textbf{PiT:} The searchable PIT transformer blocks include the base dim (ranging from $16$ to $40$, with a step of $8$), and the MLP ratio (ranging from $2$ to $8$, with a step of $2$). The number of heads can be selected from the set \{2,4,8\}.
    The depth choices for the three stages of PIT are  \{1,2,3\}, \{4,6,8\}, and \{2,4, 6\}.
\end{itemize}

% For tiny datasets, we set the parameter intervals of AutoFormer and PiT in our experiments are $4\sim9$ M and $2\sim25$ M, respectively. 

\begin{table}[t]
% \vspace{-7mm}
\caption{
Searchable components of the AutoFormer and PiT search space.
Tuples of three values in parentheses represent the lowest value, highest, and steps.
Meanwhile, values in curly brackets represent a set of all possible options.
}
\centering
\setlength{\tabcolsep}{2pt}
\resizebox{\columnwidth}{!}{
\begin{tabular}{l|c|l|c}
\toprule
\multicolumn{2}{c|}{AutoFormer}   &      \multicolumn{2}{c}{ PiT}  \\ \midrule
Embed Dim        & (192, 240, 24)       & Base Dim   & (16, 40, 8)   \\
$Q$-$K$-$V$ Dim         & (192, 256, 64)       & Patch Size  & 16      \\
MLP Ratio  & (3.5, 4, 0.5)   & MLP Ratio  & (2, 8, 2)   \\
Head Num   &  (3, 4, 1)    & Head Num  & \{2,4,8\}   \\
Depth Num         & (12, 14, 1)      & Depth Num  & \{\{1,2,3\},\{4,6,8\},\{2,4,6\}\}  \\
\midrule
Params Range & 4 -- 9M & Params Range & 2 -- 25M  \\
\bottomrule
\end{tabular}}
% \vspace{-4mm}
\label{tab:Search Space}
\end{table}

\subsection{Training settings}

\subsubsection{Vanilla Training settings}
Vanilla training refers to a standard training procedure in which a ViT model learns directly from labeled data, with no external guidance or prior knowledge. The loss function in this approach typically measures the discrepancy between predicted and true labels, with the optimization process seeking to minimize this loss. For all ViT models, we follow the training settings described in \cite{liu2021efficient}. More specifically, we use the AdamW optimizer \cite{Loshchilov2017DecoupledWD} with an initial learning rate of 5e-4 and a weight decay of 0.05. We employ a cosine policy \cite{loshchilov2016sgdr} for the learning rate schedule, eventually reducing the learning rate to 5e-6. Each ViT is trained for a total of 300 epochs, with a linear warm-up period of 20 epochs, using a batch size of 128. We train on images with an interpolated resolution of $224\times224$.

\subsubsection{Distillation Training settings}
Distillation training is a specialized training approach in which a student model (in this case, the ViT) is trained under the guidance of a pre-trained teacher model.  
During distillation training, the loss function incorporates both the standard supervised loss, based on the labeled data, and a distillation loss, based on the teacher model's predictions.
We adopt the distillation technique proposed in \cite{li2022locality}. This method utilizes a lightweight CNN to provide local information to significantly improve the performance of the ViTs, particularly on small datasets.
The training hyper-parameters for ViTs are the same as those used in vanilla training.
When training the CNN teacher (ResNet56), we use the SGD optimizer with an initial learning rate of 0.1 and a weight decay of 5e-4.
The CNN teacher takes low-resolution images ($32\times 32$) as inputs, which facilitates efficient training.
The hyper-parameters and training strategies for both ViTs and ResNet56 are summarized in Table \ref{table:train-setting}.

\begin{table}[t]
% \vspace{-7mm}
\caption{
The training hyper-parameters on datasets including CIFAR-100, Oxford Flowers, and Chaoyang.
}
\centering
\setlength{\tabcolsep}{2pt}
\begin{tabular}{l|c|l|c}
\toprule
\multicolumn{2}{c|}{ViT}   &      \multicolumn{2}{c}{ ResNet56}  \\ \midrule
optimizer        & AdamW       & optimizer  & SGD    \\
input size         & $224\times224$       & input size  & $32\times32$      \\
initial lr  & 5e-4   & base lr  & 0.1   \\
ending lr   & 5e-6    & momentum  & 0.9   \\
warm-up         & 20      & nesterov  & True   \\
weight decay     & 0.05     & weight decay  & 5e-4   \\
batch size       & 128       & batch size & 128   \\
epoch       & 300   & epoch  & 300   \\
LR schedule  & cosine   & LR schedule  & cosine   \\
\bottomrule
\end{tabular}
% \vspace{-4mm}
\label{table:train-setting}
\end{table}

\begin{table}
\caption{
ViT-Bench-101 provides the following metrics.
`Dis Acc' means accuracy obtained with distillation.
}
\setlength{\tabcolsep}{1pt}
\centering
\begin{tabular}{c|c|c|c}
\toprule
Search Space    &  Num   & Dataset    & Metric \\\midrule
\multirow{4}{*} {AutoFormer}     &\multirow{4}{*} {500}     & CIFAR-100             & Dis Acc, Vanilla Acc        \\
   &   & Flowers       & Dis Acc, Vanilla Acc          \\
   &    & Chaoyang             & Dis Acc, Vanilla Acc          \\
   &    & ImageNet-1K            & Subnet Acc          \\ \midrule
\multirow{3}{*} {PiT}  & \multirow{3}{*} {500}  & CIFAR-100              & Dis Acc, Vanilla Acc           \\
  & & Flowers              &  Dis Acc, Vanilla Acc          \\
&  & Chaoyang             & Dis Acc, Vanilla Acc            \\
\bottomrule
\end{tabular}
% \vspace{-6mm}
\label{table:metric}
\end{table}

\subsection{Metrics}

ViT-Bench-101 delivers different evaluation metrics for assessing the performance of ViTs. 
Specifically, we provide two types of accuracy metrics on small datasets: distillation accuracy and vanilla accuracy. 
Distillation accuracy reflects the performance of a ViT when it is trained using the knowledge distillation technique, while vanilla accuracy indicates the performance of a ViT when it is trained from scratch without any form of teacher guidance.
Additionally, we evaluate the accuracy of sub-networks sampled from the AutoFormer supernet on the ImageNet-1K dataset. 
As demonstrated by the AutoFormer study, these sampled subnets from the supernet can achieve performance comparable to that of the same models trained from scratch.
We provide a comprehensive list of supported metrics for different datasets in Table~\ref{table:metric}, allowing users to easily select the appropriate metrics for their specific evaluation needs.

\subsection{The usage of API} 

% To make it easier for users to access and utilize our benchmark, we provide an API that allows users to quickly query the results of each trial for the specified metric 'A'. 
% This feature is particularly useful for researchers and practitioners who need to compare and analyze different models, as it allows them to retrieve the information they need without incurring significant computational costs.

We provide a convenient API to make it easier for users to access and utilize our ViT-Bench-101 benchmark.
The dataset can be easily installed using the command "pip install -e ."
” in our Auto-Prox repository.
The following code snippet shows how to use the ViT-Bench-101 dataset: 

\begin{lstlisting}[language=Python]
from Auto-Prox.api import API 
api = API('AutoFormer-GT.pkl', verbose=False)
# sample random index 
rnd_idx = api.random_index()

# query arch_cfg by index 
arch_cfg = api.arch_by_idx(rnd_idx)
print(f'The index: {rnd_idx} arch_cfg: {arch_cfg}')

# query accuracy by index 
acc = api.acc_by_idx(rnd_idx, dataset, distill = True)
print(f'Acc on {dataset}: {acc:.4f}')


\end{lstlisting}

% As presented in the usage, our ViT-Bench-101 provides vanilla and knowledge distillation accuracies for multiple datasets. 
% In knowledge distillation, the provided teachers include resnet32, resnet56, and resnet110.
We will release the code and benchmark data file.

% \begin{itemize}
%     \item The unary operations available in the search space include ``log", ``abslog", ``abs", ``pow", ``exp", ``normalize", ``relu", ``swish", ``mish", ``leaky\_relu", ``tanh", ``invert", ``frobenius\_norm", ``normalized\_sum", ``l1\_norm", ``softmax", ``sigmoid", ``logsoftmax", ``sqrt", ``revert", ``min\_max\_normalize", ``to\_mean\_scalar", ``to\_std\_scalar", and ``no\_op." 
%     % These operations can be applied to a single input tensor, which may have any number of dimensions.
%     \item The binary operations available in the search space are ``sum", ``subtract", ``multiply", and ``dot." 
%     % These operations can be applied to two input tensors, which may have any number of dimensions, as long as they are compatible for the specified operation.
% \end{itemize}

% Specifically, the available unary operations include ”log,” ”abslog,” ”abs,” ”pow,” ”exp,” ”normalize,” ”relu,” ”invert,” ”frobenius norm,” ”normalized sum,” ”l1 norm,” ”softmax,” ”sigmoid,” ”logsoftmax,”
% ”sqrt,” ”revert,” ”min max normalize,” ”to mean scalar,”
% ”to std scalar,” and ” no-op.” 
% Comparatively, the binary operations are
% ”sum,” ”subtract,” ”multiply,” and ”dot.” 
% Their detailed descriptions are presented in Table~\ref{tab:operation}.  

\section{Ranking correlation metrics}
The ranking correlation metric is a statistical measure that evaluates the association between the rankings of two variables. 
It allows researchers to evaluate how well the relative order of values in a zero-cost proxy matches the relative order of values in the ground-truth dataset. 
The higher the ranking correlation metric, the more accurately the zero-cost proxy reflects the actual accuracy in rank order.

We use $\beta_i$ to represent the ground-truth (GT) accuracy of the Vision Transformer (ViT) $\alpha_i$, where $i$ falls within the range ${1,...,N}$. Similarly, we denote its zero-cost proxy score as $\gamma_i$. The rankings of the GT and estimated proxy scores are represented by $r_i$ and $k_i$, respectively. We employ three correlation metrics in our analysis: the Pearson coefficient ($r$), Kendall's Tau ($\tau$), and Spearman coefficient ($\rho$). The calculations for these metrics are as follows:

\begin{itemize}
	\item Pearson correlation coefficient (Linear Correlation):
	      $r=\mbox{corr}(\beta, \gamma)/\sqrt{\mbox{corr}(\beta, \beta) \mbox{corr}(\gamma, \gamma)}$.
	\item Kendall's Tau correlation coefficient: The relative difference of concordant pairs and discordant pairs $\tau=\sum_{i < j} \text{sgn}(\beta_i - \beta_j)\text{sgn}(\gamma_i - \gamma_j) / {M \choose 2}$.
	\item Spearman correlation coefficient: The Pearson correlation coefficient between the ranking variables $\rho=\mbox{corr}(r, k) / \sqrt{\mbox{corr}(r, r)\mbox{corr}(k, k)}$.
\end{itemize}

Pearson evaluates the linear correlation between two variables, whereas both Kendall's Tau and Spearman assess the monotonic relationship. These metrics yield values ranging from -1 to 1: -1 signifies a negative correlation, 1 denotes a positive correlation, and 0 indicates no relationship.

\begin{figure}[t]
    \centering
    \includegraphics[width=\linewidth]{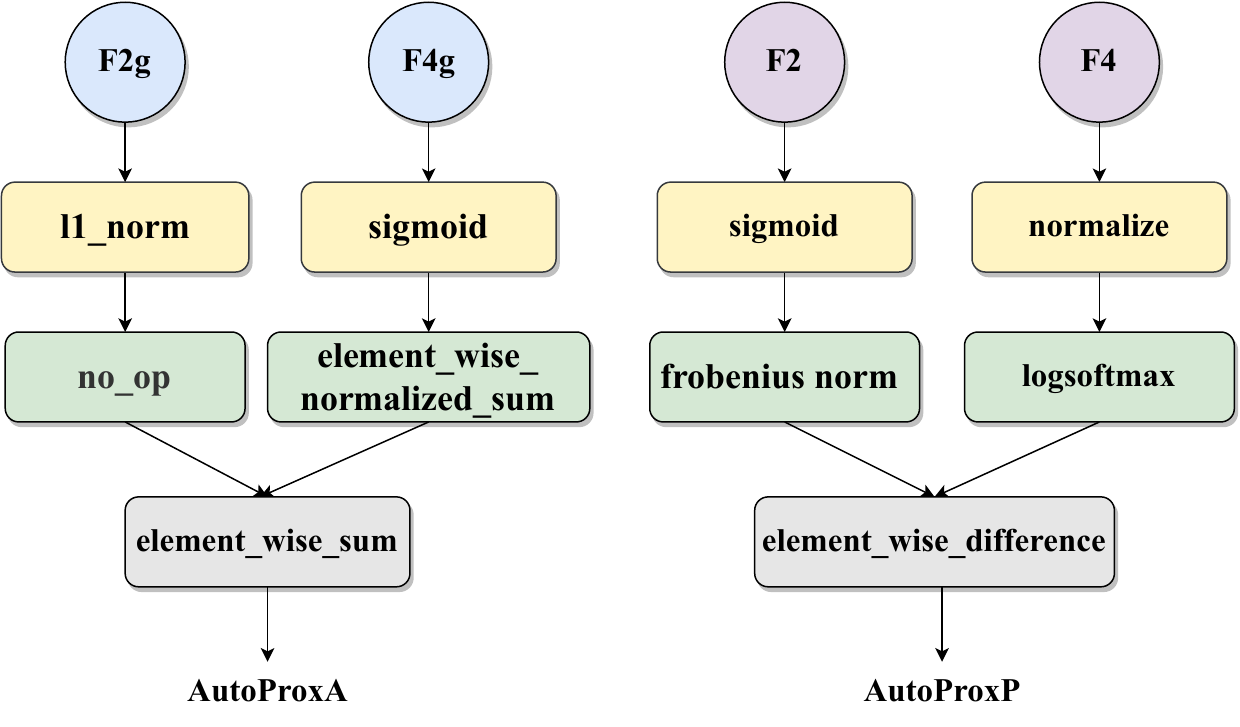}
    % \vspace{-1em}
    \caption{Illustration of searched zero-cost proxies on AutoFormer (Left) and PiT (Right) search spaces.
    % using activation (Act), gradient (Grad), and weights (Weight) with Linear layers of self-attention and feed-forward networks as inputs.
    }
    \label{fig:searched_porxy}
\end{figure}

\section{Computation Graph Details}

In \ourmethod{}, each proxy within the population is encoded as a computation graph, comprising two inputs, a series of primitive operations, and a single output. For each input, there are eight possible selections (F1, F1g, F2, F2g, F3, F3g, F4, F4g). Additionally, there are 24 potential unary operations for processing each input, followed by four available binary operations to generate an output. The computation graph is designed to accept two types of network statistics as inputs, which can be either identical or distinct. For each branch of the computation graph, we apply two consecutive unary operations to transform the network statistics. 
Subsequently, we use a binary operation that takes the outputs of the two branches as input and applies the $\textit{to\_mean\_scalar}$  aggregation function to produce the final output. 
Figure~\ref{fig:searched_porxy} showcases the searched zero-cost proxies on Vision Transformers (ViTs) sampled from the AutoFormer and PiT search spaces.

\section{Validity Check for Zero-cost Proxies}
% During the proxy search process, the validity of the newly generated zero-cost proxies is crucial. 

We conduct a Validity Check for newly generated zero-cost proxies to reduce the number of invalid proxies during the search process. 
Specifically, we check if the proxy score of a proxy belongs to a set of invalid scores, which includes \{-1, 1, 0, NaN, and Inf\}. 
These scores indicate that the proxy is invalid and inapplicable.

Invalid proxy scores may manifest in several ways. 
For instance, a score of -1 could result from shape mismatch issues or user-defined exceptions, while a score of 1 may indicate a numerical insensitivity. 
A score of 0 may indicate a numerical instability or an invalid operation. 
The occurrence of NaN can be attributed to various factors, including division by zero, taking the square root of a negative number, or computing the logarithm of a non-positive number. 
A score of infinity (Inf) could be a result of overflow in arithmetic operations or arise from invalid mathematical operations.

We identify and filter out these invalid proxies after mutation, reducing the sparsity of the search space when seeking effective zero-cost proxies.

\section{More Experiments and Ablation Studies}

\begin{table}[t]
    \centering
    \label{tab:OnDifferentSeeds}
       \caption{The effect of different seeds on the Kendall correlation results (\%) of various zero-cost proxies.
    ViTs-accuracy configurations are sampled from the PiT search space and CIFAR-100 of ViT-Bench-101.
           % For each random seed, we sample 100 vision transformers.
    % For each random seed, we sample 50 vision transformers, then calculate Kendall correlation results from their distillation ground truths and zero-proxy scores.
    }
    \small{
     % \resizebox{1\linewidth}{!}{
    \begin{tabular}{l|ccc|cc}
    \toprule
    \multirow{2}{*}{Proxy}      & \multicolumn{3}{c|}{random seed} & \multirow{2}{*}{AVG} & \multirow{2}{*}{STD} \\
    \cline{2-4}
        & 0 & 1 & 2  & \\
    \midrule
    % SNIP        &  &  &  & &  &   \\
    GraSP       &-49.32  & -45.17 & -31.56  & -42.02    & 0.58 \\
     SynFlow       & 77.06 & 77.72 & 54.59     & 69.79 &  1.16\\
    TE-NAS   & 4.08 &  -1.31  & -9.17  & -2.13   & 0.30 \\
    NWOT      & -2.12  & -2.18 &  -3.54 & -2.61     & 0.01  \\
    TF-TAS   & 67.35 & 62.03  & 62.12 & 63.83   & 0.06  \\
        \midrule
        \ourmethod{} (Ours) & 86.19 & 86.04  & 73.98    &  \textbf{82.07} & 0.33 \\
        \bottomrule
    \end{tabular}
    }
    % }
    % \vspace{-4mm}
    \label{tab:robust}
\end{table}

% \subsection{Robust experiment}
\subsection{Sensitivity Analysis of Seeds}
To check the stability of various zero-cost proxies, we conduct experiments using three different seeds and compute the average and standard deviation.
For each seed, we randomly sample 100 ViTs, then calculate the Kendall Tau correlation (\%) between their distillation accuracy and corresponding zero-cost proxy scores.
As shown in Table~\ref{tab:robust}, \ourmethod{} consistently achieves superior performance to other zero-cost proxies.
These findings underscore the enhanced stability and reliability of our method in comparison to other alternative approaches.

\subsection{Analysis of Iterations}
\begin{figure}[t]
	\setlength{\abovecaptionskip}{0.cm}
	\setlength{\belowcaptionskip}{-0.cm}
	\centering
	\begin{minipage}[t]{0.49\linewidth}
		\centering
		\includegraphics[width=1\linewidth]{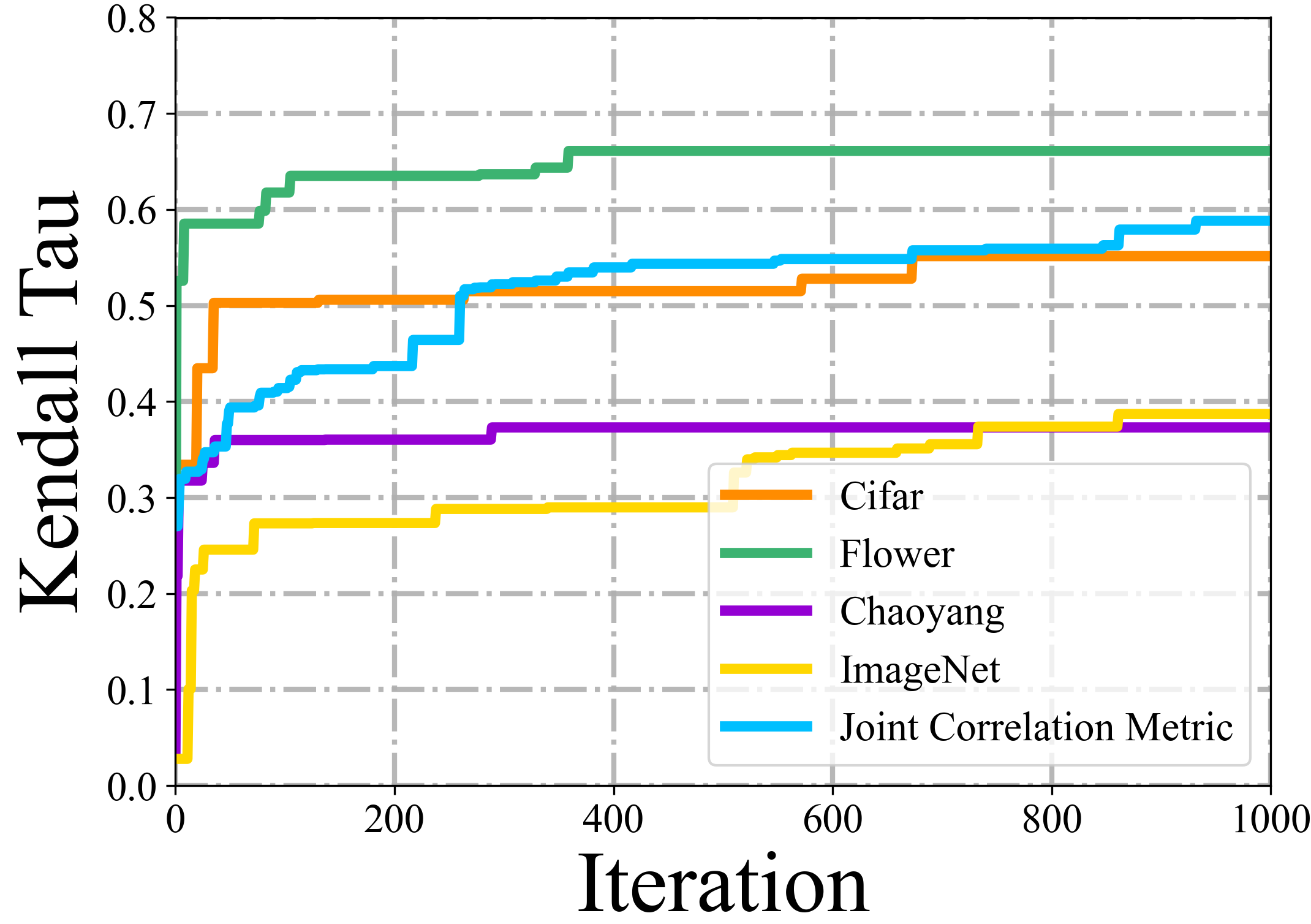}
	\end{minipage}
	\begin{minipage}[t]{0.49\linewidth}
		\centering
		\centering
		\includegraphics[width=1\linewidth]{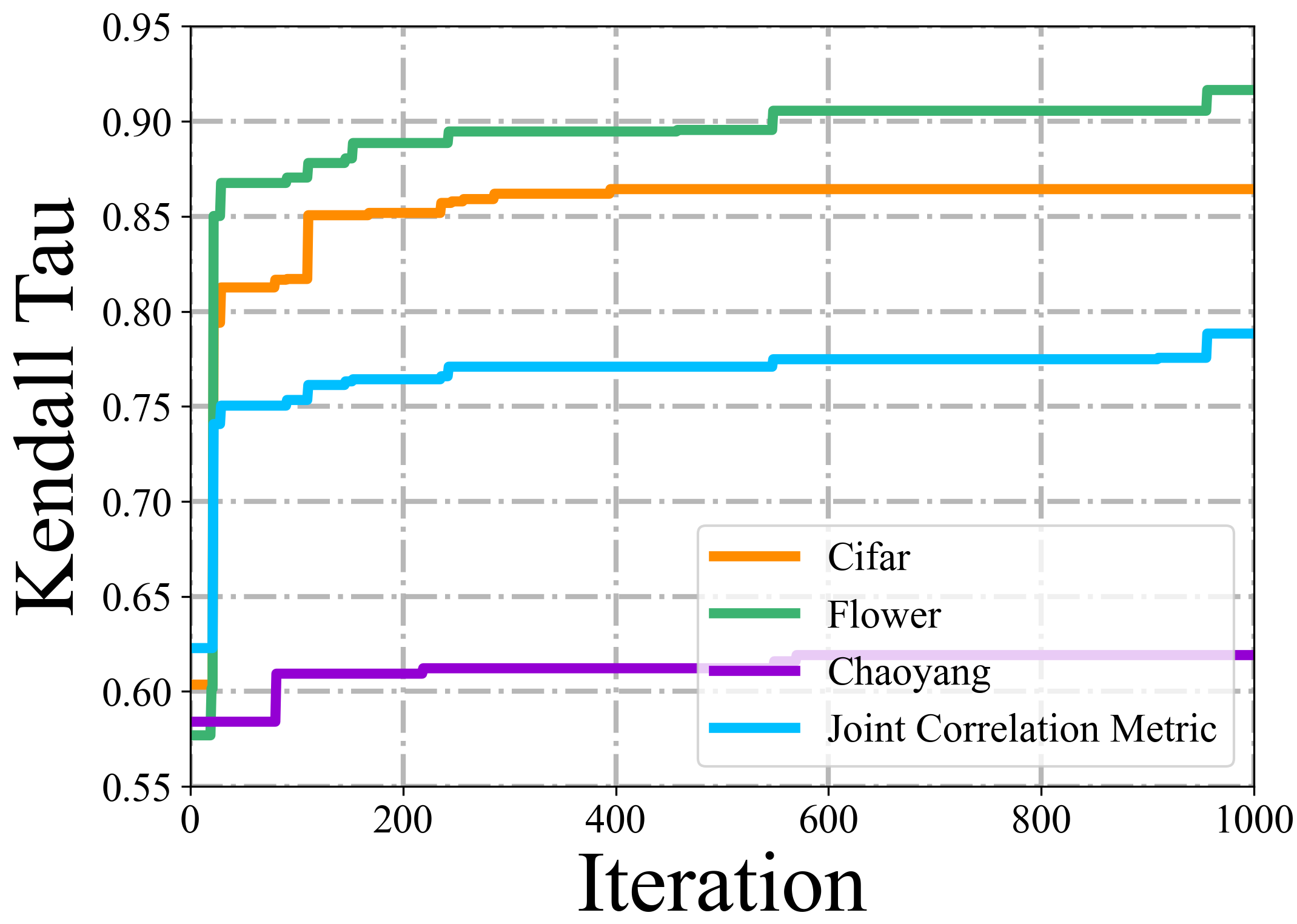}
	\end{minipage}\label{fig:structure}
	\caption{
 The 1000-iteration evolutionary processes of Kendall Tau on each dataset and the joint correlation metric.
 We search proxies for ViTs sampled from the AutoFormer search space (Left) and the PiT search space (Right).
 }
\label{fig:evo_1000}
 % \vspace{-1em}
\end{figure}

In Figure \ref{fig:evo_1000}, we provide an expanded view of the results from our evolutionary proxy search over a larger number of iterations. 
It can be observed that in the first 200 iterations, the evolutionary curve showed a clearly increasing trend.
However, as the number of iterations continued to increase, only marginal improvements were achieved.
Given these observations, and in pursuit of both computational efficiency and optimal performance, we have chosen to set the default number of evolutionary iterations at 200.

% We present more iterations of the evolutionary proxy search in Figure \ref{fig:evo_1000}.

\subsection{Training-free Search for Larger ViTs }
We utilized the searched zero-cost proxy, denoted as AutoProxA, to conduct a training-free search for ViT models on AutoFormer-B~\cite{chen2021autoformer}, within which the ViTs span a parameter range of approximately $42\sim 75$ M. 
Our goal is to explore the applicability and generality of our method for larger ViT models.
In our experiments, we searched for 200 ViTs using different zero-cost proxies and reported the final distillation accuracy, model parameters, and search costs. 
As depicted in Table~\ref{tab:large_search_space}, our method can achieve even better results, suggesting that our method has the ability to adapt and generalize effectively to larger ViT models.

% \subsection{More Visualization of Ranking Correlation}

\begin{table}[t]
	\centering
  % \vspace{-2mm}
  	\caption{Training-free ViT search results using various zero-cost proxies in the AutoFormer-B search space. We report distillation accuracy on the CIFAR-100 dataset.}
 \label{tab:large_search_space}
     % \vspace{-3mm}
	%\setlength{\tabcolsep}{3.3pt}
	% \footnotesize
	 \resizebox{1\linewidth}{!}{
	\begin{tabular}{c|c|ccc}
		\toprule
		Search Space           & Proxy  & Param (M)& Search Cost        &  Dis. Acc(\%)                                                        \\
		\midrule
		% \multirow{4}{*} {AutoFormer-S}        & GraSP               &  78.88    & 35.06                 &       0.68                                          \\
  %            & synflow               &  79.24       & 30.80                 &       0.50                                         \\
  %             & DSS               &  79.69    & 30.03                 &       0.50                                          \\
  %              & TVT               &  \textbf{79.94}    & 30.46                 &       \textbf{0.42}                                          \\
		% \midrule
			\multirow{4}{*} {AutoFormer-B}        & GraSP                  &   68.94               &   1.79 h    &  80.22                                           \\
             & SynFlow                   &   70.91              & 1.06 h                 &  80.33                             \\
     & TF-TAS                  & 59.86                &  0.97 h       &  80.28                                        \\
               & Ours                  &  \textbf{52.22}                &  \textbf{0.55 } h             & \textbf{80.51}                                  \\
  %   \multirow{4}{*} {Training-free} 
  % %       & NWOT~\cite{NWOT}                 &                         &                         &                             \\
		% %                                
  %           & TE-NAS                 &   75.84                      &   1.01                                  \\
  %        & DSS                 &     77.78                    &   0.08                                           \\   
		\bottomrule
	\end{tabular}}
 % \vspace{-4mm}
\end{table}

\subsection{Generalize to Other Architecture Search Space}
To validate the generalization of \ourmethod{} proxy to other architecture search space, we test the discovered AutoProxP on NAS-Bench-201~\cite{nas201}.
NAS-Bench-201 is a popular benchmark dataset that contains 15k unique CNN architectures.
The results in Table~\ref{tab:nasbench201} show that AutoProxP exhibits strong rank consistency,  indicating its ability to generalize well to other architecture search spaces.
We compare EZNAS-A~\cite{akhauri2022eznas} and AutoProxP in NAS-Bench-201. The results in Table~\ref{tab:nasbench201} show that AutoProxP exhibits superior ranking consistency.

\begin{table}[t]
\centering
\caption{Ranking correlation (\%) on NAS-Bench-201 CIFAR-10. 
$\,{\dagger}$ means our implementation.
}
\label{tab:nasbench201}
% \vspace{-3mm}
\resizebox{0.75\linewidth}{!}{
\begin{tabular}{cccc}
\hline
Proxy   & Kendall  & Spearman & Pearson \\ \hline
Snip & 51.16 & 69.12  &  28.20        \\
SynFlow & 58.83 & 77.64  &  44.90        \\
NWOT  & 55.13 &  74.93    & 84.18   \\
EZNAS-A &   65.00   & 83.00     & -    \\
EZNAS-A$\,^{\dagger}$    &  64.21    & 83.16      &  84.99 \\
AutoProxP & \textbf{68.49} & \textbf{84.81} & \textbf{86.71}   \\ \hline
\end{tabular}
}
% \vspace{-2mm}
\end{table}

\begin{table*}[t]
\small
 \centering
  	\caption{Searched architectures with \ourmethod{}. 
 }
\begin{tabular}{cc}
\toprule
\multicolumn{1}{c|}{}           & Searched architectures                                                                                                                  \\ \midrule
\multicolumn{2}{c}{CIFAR-100}                                                                                                                                             \\ \midrule
\multicolumn{1}{c|}{AutoFormer} & \tabincell{c}{\{'mlp\_ratio': [3.5, 4.0, 3.5, 4.0, 3.5, 3.5, 4.0, 4.0, 4.0, 4.0, 4.0, 3.5, 4.0, 3.5], \\
'num\_heads': [4, 3, 3, 3, 4, 4, 4, 3, 4, 4, 4, 4, 3, 4], 'hidden\_dim': 240, 'depth': 14\}} \\ \midrule
\multicolumn{1}{c|}{PIT}        & \tabincell{c}{\{'mlp\_ratio': 6, 'num\_heads': [2, 2, 8], 'base\_dim': 32, 'depth': [3, 4, 6]\}}    \\ \midrule
\multicolumn{2}{c}{Flower}                                                                                                                                                \\ \midrule
\multicolumn{1}{c|}{AutoFormer} & \tabincell{c}{\{'mlp\_ratio': [3.5, 4.0, 4.0, 4.0, 4.0, 3.5, 4.0, 4.0, 4.0, 4.0, 4.0, 3.5, 4.0, 4.0], \\
'num\_heads': [4, 4, 3, 4, 4, 4, 3, 3, 4, 4, 4, 3, 3, 3], 'hidden\_dim': 240, 'depth': 14\}}
 \\ \midrule
\multicolumn{1}{c|}{PIT}        & \tabincell{c}{\{'mlp\_ratio': 8, 'num\_heads': [4, 4, 8], 'base\_dim': 32, 'depth': [2, 8, 2]\}}   \\ \midrule
\multicolumn{2}{c}{Chaoyang}                                                                                                                                              \\ \midrule
\multicolumn{1}{c|}{AutoFormer} & \tabincell{c}{\{'mlp\_ratio': [4.0, 4.0, 3.5, 4.0, 3.5, 4.0, 3.5, 4.0, 4.0, 4.0, 4.0, 3.5, 3.5], \\
'num\_heads': [3, 4, 4, 4, 3, 3, 3, 3, 4, 3, 4, 3, 3], 'hidden\_dim': 240, 'depth': 13\}} \\ \midrule
\multicolumn{1}{c|}{PIT}        & \tabincell{c}{\{'base\_dim':40, 'num\_heads': [2, 4, 4], 'mlp\_ratio': 6, 'depth': [2, 4, 6]\}}   \\ 
\midrule
\multicolumn{2}{c}{ImageNet}                                                                                                                                              \\ \midrule
\multicolumn{1}{c|}{AutoFormer} & \tabincell{c}{\{'mlp\_ratio': [3.5, 4.0, 4.0, 3.5, 3.5, 3.5, 3.5, 4.0, 4.0, 3.5, 3.5, 4.0, 3.5], \\ 'num\_heads': [4, 4, 3, 4, 4, 3, 4, 4, 4, 3, 4, 3, 4], 'hidden\_dim': 192, 'depth': 13\}} \\
\bottomrule
\end{tabular}
 \label{tab:searched_results}
\end{table*}

\begin{table*}
    \centering
    \caption{The unary and binary operations in the proxy search space. "UOP" denotes the unary operation, and "BOP" denotes the binary operation.}
    \label{tab:operation}
    \begin{tabular}{l|llll}
        \toprule[1.1pt]
        OP ID & OP Name                                  & Input Args        & Output Args   & Description                                         \\    \midrule[0.8pt]
        UOP00   & \textcode{no\_op}                        & --                & --            & --                                                  \\ \midrule[0.8pt]
        UOP01 & \textcode{element\_wise\_abs}            & $a$ / scalar,matrix     & $b $ / scalar,matrix & $x_b=\left|x_a\right|$                              \\
        UOP02 & \textcode{element\_wise\_tanh}            & $a$ / scalar,matrix     & $b $ / scalar,matrix & $x_b=\tanh(x_a)$                                          \\
        UOP03 & \textcode{element\_wise\_pow}            & $a$ / scalar,matrix & $b$ / scalar,matrix & $x_b=x_a^2$                                       \\
        UOP04 & \textcode{element\_wise\_exp}            & $a$ / scalar,matrix     & $b $ / scalar,matrix & $x_b=e^{x_a}$                                       \\
        UOP05 & \textcode{element\_wise\_log}            & $a$ / scalar,matrix     & $b $ / scalar,matrix & $x_b=\ln x_a$                                       \\
        UOP06 & \textcode{element\_wise\_relu}           & $a$ / scalar,matrix     & $b $ / scalar,matrix & $x_b=\max(0,x_a)$                                   \\
        UOP07 & \textcode{element\_wise\_leaky\_relu}           & $a$ / scalar,matrix     & $b $ / scalar,matrix & $x_b=\max(0.1x_a,x_a)$                                   \\
        UOP08 & \textcode{element\_wise\_swish}           & $a$ / scalar,matrix     & $b $ / scalar,matrix & $x_b=x_a\times \text{sigmoid}(x_a)$                                   \\
        UOP09 & \textcode{element\_wise\_mish}           & $a$ / scalar,matrix     & $b $ / scalar,matrix & $x_b=x_a\times \text{tanh}(\ln{1+\text{exp}(x_a)})$                                   \\
        UOP10 & \textcode{element\_wise\_invert}         & $a$ / scalar,matrix     & $b $ / scalar,matrix & $x_b=1/x_a$                                         \\
        UOP11 & \textcode{element\_wise\_normalized\_sum} & $a$ / scalar,matrix & $b$ / scalar,matrix & $x_b=\frac{\sum x_a}{\text{numel}(x_a)+\epsilon}$                     \\
        UOP12 & \textcode{normalize}                     & $a$ / scalar,matrix     & $b $ / scalar,matrix & $x_b=\frac{x_a - \text{mean}(x_a)}{\text{std}(x_a)}$                    \\
        UOP13 & \textcode{sigmoid}                       & $a$ / scalar,matrix     & $b $ / scalar,matrix & $x_b=\frac{1}{1+e^{-x_a}}$                          \\
        UOP14 & \textcode{logsoftmax}                    & $a$ / scalar,matrix     & $b $ / scalar,matrix & $x_b=\ln\frac{e^{x_a}}{\sum_{i=1}^n e^{s_i}}$       \\
        UOP15 & \textcode{softmax}                       & $a$ / scalar,matrix     & $b $ / scalar,matrix & $x_b=\frac{e^{x_a}}{\sum_{i=1}^n e^{s_i}}$          \\
        UOP16 & \textcode{element\_wise\_sqrt}           & $a$ / scalar,matrix     & $b $ / scalar,matrix & $x_b=\sqrt{x_a}$                                    \\
        UOP17 & \textcode{element\_wise\_revert}         & $a$ / scalar,matrix     & $b $ / scalar,matrix & $x_b=-x_a$                                         \\
        UOP18 & \textcode{frobenius\_norm}                & $a$ / scalar,matrix     & $b $ / scalar,matrix & $x_b=\sqrt{\sum_{i=1}^n s_i^2}$                     \\
        UOP19 & \textcode{element\_wise\_abslog}         & $a$ / scalar,matrix     & $b $ / scalar,matrix & $x_b=\left|\ln x_a\right|$                          \\
        UOP20 & \textcode{l1\_norm}                      & $a$ / scalar,matrix     & $b $ / scalar,matrix & $x_b=\frac{\sum_{i=1}^n \left|s_i\right|}{\text{numel}(x_a)}$                 \\
        UOP21 & \textcode{min\_max\_normalize}           & $a$ / scalar,matrix     & $b $ / scalar,matrix & $x_b=\frac{x_a-\min(x_a)}{\max(x_a)-\min(x_a)}$     \\
        UOP22 & \textcode{to\_mean\_scalar}              & $a$ / scalar,matrix     & $b $ / scalar & $x_b=\frac{x_a}{n}$                                 \\
        UOP23 & \textcode{to\_std\_scalar}               & $a$ / scalar,matrix     & $b $ / scalar & $x_b=\sqrt{\frac{\sum_{i=1}^n (s_i-\bar{s})^2}{n}}$ \\ \midrule[0.8pt]

        BOP01 & \textcode{element\_wise\_sum}            & $a$,$b$ / scalar,matrixs & $c $ / scalar,matrix & $x_c=x_a+x_b$                                       \\
        BOP02 & \textcode{element\_wise\_difference}     & $a$,$b$ / scalar,matrixs & $c $ / scalar,matrix & $x_c=x_a-x_b$                                       \\
        BOP03 & \textcode{element\_wise\_product}        & $a$,$b$ / scalar,matrixs & $c $ / scalar,matrix & $x_c=x_a\times x_b$                                       \\
        BOP04 & \textcode{matrix\_multiplication}        & $a$,$b$ / scalar,matrixs & $c $ / scalar,matrix & $x_c=x_a @x_b$                                     \\ \bottomrule[1.1pt]
    \end{tabular}
\end{table*}

% \section{More experimental details and results}

\section{Searched ViT with \ourmethod{}}
We provide details of searched architectures in the AutoFormer and PiT search spaces. 
These results can be found in Table \ref{tab:searched_results}. 
In the AutoFormer search space, the terms "mlp\_ratio" and "num\_heads" correspond to the values in all transformer layers.
The "hidden\_dim" refers to the embedding dimension, while "depth" denotes the number of transformer layers.
Differently, the architectures within the PiT search space are stage-based.
Both "num\_heads" and "depth" have three values, each representing the result from the three distinct stages.

\section{Primitive Operations}

The primitive operations utilized in \ourmethod{} are classified into two categories based on their inputs.
The unary operations are applied to a single input tensor, while the binary operations are conducted on two input tensors.
The search space of the automatic proxy consists of 24 unary operations and 4 binary operations.
Their detailed formulations are presented in Table~\ref{tab:operation}.

\bibliography{aaai24}

\end{document}